\documentclass[10pt,journal,compsoc]{IEEEtran}
\usepackage[tight,footnotesize]{subfigure}
\usepackage{graphics}
\usepackage{graphicx}
\usepackage[cmex10]{amsmath}
\usepackage{multirow,tabularx}
\usepackage{booktabs}
\usepackage{algorithm}
\usepackage{algorithmic}
\usepackage{stfloats}

\ifCLASSOPTIONcompsoc
  \usepackage[nocompress]{cite}
\else
  \usepackage{cite}
\fi
\ifCLASSINFOpdf
\else
\fi
\begin{document}
%
\title{Comprehensive Feature-based Robust Video Fingerprinting Using Tensor Model}
%
%
%
%

\author{Xiushan~Nie,
        Yilong~Yin,
        and~Jiande~Sun
\IEEEcompsocitemizethanks{\IEEEcompsocthanksitem X. S. Nie was with the School
of Computer Science and Technology, Shandong University of Finance and Economics, Jinan, 250014, China.\protect\\

\IEEEcompsocthanksitem Y. L. Yin was with the School
of Computer Science and Technology, Shandong University, Jinan,
250100, China.

\IEEEcompsocthanksitem J. D. Sun was with the School of information science and engineering, Shandong Normal university, Jinan, 250014, China.

E-mail: ylyin@sdu.edu.cn, niexsh@sdufe.edu.cn, jiandesun@hotmail.com}

}

%
%

\markboth{}%
{Shell \MakeLowercase{\textit{et al.}}: Bare Demo of IEEEtran.cls for Computer Society Journals}
%



\IEEEtitleabstractindextext{%
\begin{abstract}
Content-based near-duplicate video detection (NDVD) is essential for effective search and retrieval, and robust video fingerprinting is a good solution for NDVD. Most existing
video fingerprinting methods use a single feature or concatenating  different features to generate video fingerprints, and show a good performance under single-mode
modifications such as noise addition and blurring. However, when they suffer combined modifications, the performance is degraded to a certain extent because
such features cannot characterize the video content completely. By contrast, the assistance and consensus among different features can improve the performance
of video fingerprinting. Therefore, in the present study, we mine the assistance and consensus among different features based on tensor model, and present a new
comprehensive feature to fully use them in the proposed video fingerprinting framework. We also analyze what the comprehensive feature really is for representing
the original video. In this framework, the video is initially set as a high-order tensor that consists of different features, and the video tensor is decomposed
via the Tucker model with a solution that determines the number of components. Subsequently, the comprehensive feature is generated by the low-order
 tensor obtained from tensor decomposition. Finally, the video fingerprint is computed using this feature. A matching strategy used for narrowing
 the search is also proposed based on the core tensor. The robust video fingerprinting framework is resistant not only to single-mode modifications, but also to the combination of them.
\end{abstract}

}

\maketitle

\IEEEdisplaynontitleabstractindextext

%
\IEEEpeerreviewmaketitle

\IEEEraisesectionheading{\section{Introduction}\label{sec:introduction}}

%
%
%
%
\IEEEPARstart{W}{ith} the development of information technology, the number of digital videos available on the Web has increased explosively, and the openness of networks have made access to video contents considerably easier and cheaper. Therefore, many illegal and useless video copies or near-duplicates appear on the Web, which are generated by simple reformatting, transformations, and editing. Illegal and useless copies result in user inconvenience when surfing the Internet. A Web user may want to search for an interesting video but may end up with many near-duplicate videos with low-quality images, which is disappointing and time-consuming. In addition, most of these copies are pirated and infringe on the video producer¡¯s copyright. Therefore, the presence of massive numbers of copies imposes a strong demand for effective near-duplicate video detection (NDVD) in many applications, such as copyright enforcement, online video usage monitoring, and video database cleansing. NDVD is a broad topic that has several goals such as finding copies of brief excerpts, partial-frame copies, and near-duplicates of entire clips. In this study, we focus on the near-duplicates of entire clips without supporting partial near-duplicate detection.

Watermarking is a traditional technology used to detect copies of images or videos. It embeds imperceptible watermarks into the media to prove its authenticity. However, the watermarks embedded into the media may cause distortion. By contrast, robust fingerprinting techniques extract the most important features of the media to calculate compact digests that allow for efficient content identification without modifying the media. Therefore, robust video fingerprinting has attracted an increasing amount of attention in the field of NDVD.

Many video fingerprinting methods have been developed in recent years; global [1-6] and local feature-based [8-11] methods are two primary types. In global feature-based methods, the video is represented as a compact global feature, such as color space [2], histograms [3], and block ordinal ranking [4]. In the temporal [5] and transformation-based methods, 3D-discrete cosine transform [6] and nonnegative matrix factorization [7], for example, can also be considered as global feature-based approaches [8], whereas the local feature-based methods use the descriptors of the local region around interest points, such as Harris interest point detector [9], scale-invariant feature transform (SIFT) [10], centroid of gradient orientations (CGO) [11], and speed-up robust feature (SURF) [12]. Global features can achieve fast retrieval speed but are less effective in handling video copies with layers of editing, such as caption/logo insertion, letter-box, and shifting. Local features are more effective for most video editing; however, they usually require more computation. Recently, Li et al. [13] used graph model embedding to generate video fingerprints. Almeida et al. [14] calculated horizontal and vertical coefficients of DCT from video frames as the video fingerprint. Sun et al. [15] proposed a temporally visual weighting method based on visual attention to produce video fingerprint. Nie et al. [16] used graph and video tomography [17] to compute video fingerprint. Although these methods used different features, they can be roughly arranged into global or local feature-based methods.

\begin{figure}[htb!]
\begin{center}
  \subfigure[] {\scalebox{0.2}{\includegraphics[width=\textwidth,height=5in]{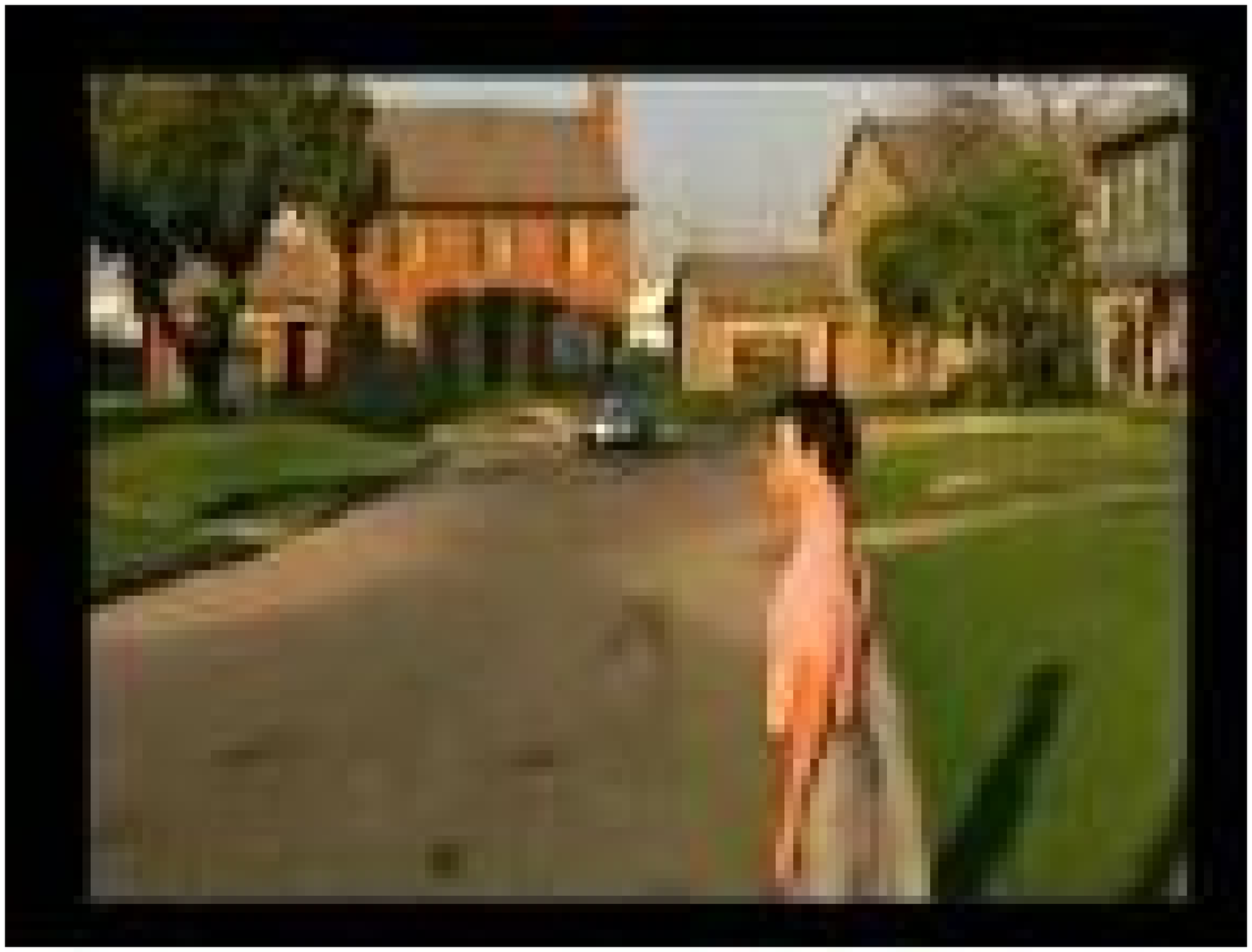}}}
  \subfigure[] {\scalebox{0.2}{\includegraphics[width=\textwidth,height=5in]{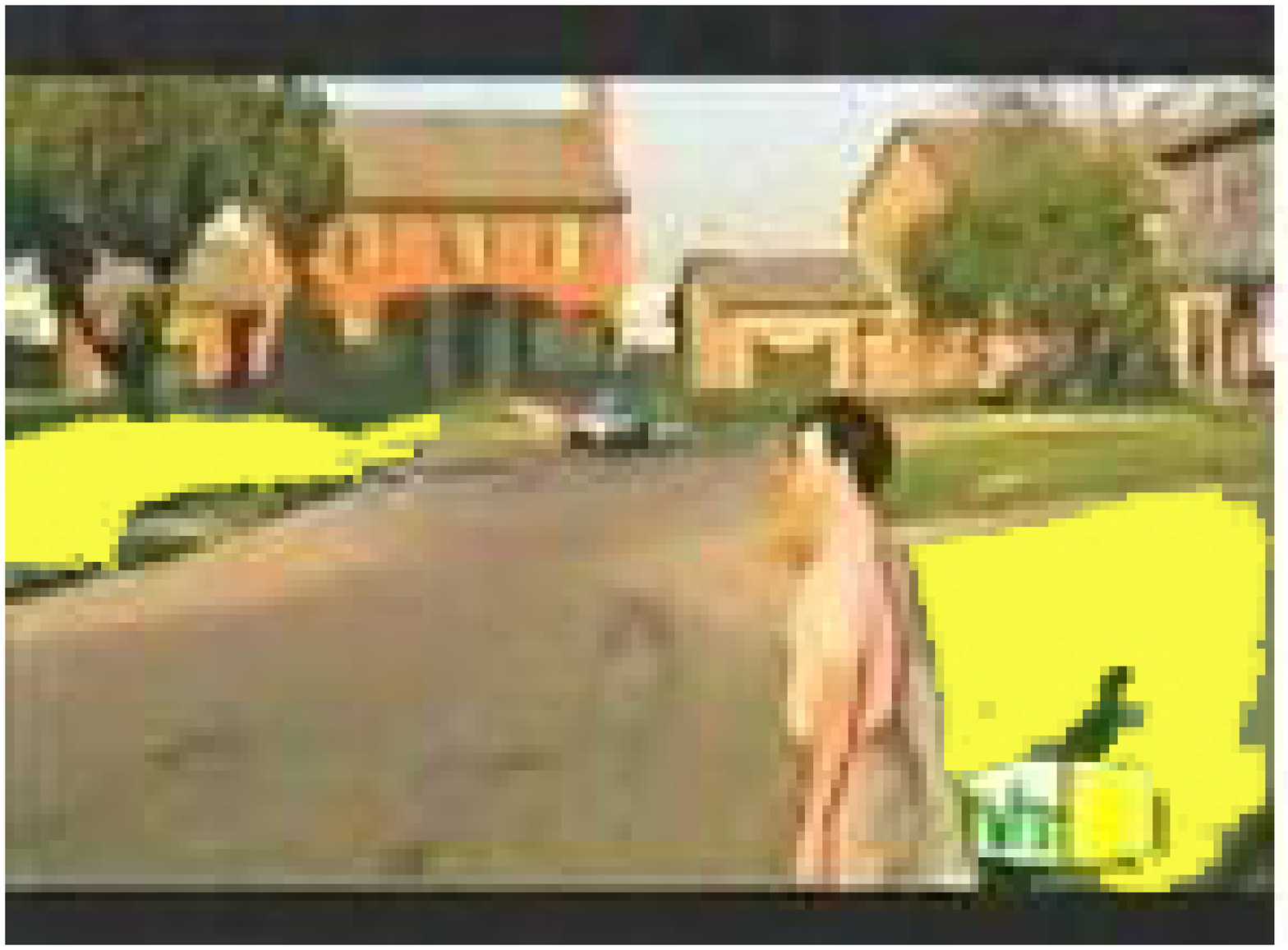}}}
  \caption{ Original picture and its copy : (a)Original; (b) Copy}\label{inter_compare}
\end{center}
\end{figure}
\begin{figure*}[htbp] \centering
\includegraphics [width=6in,height=2.7in]{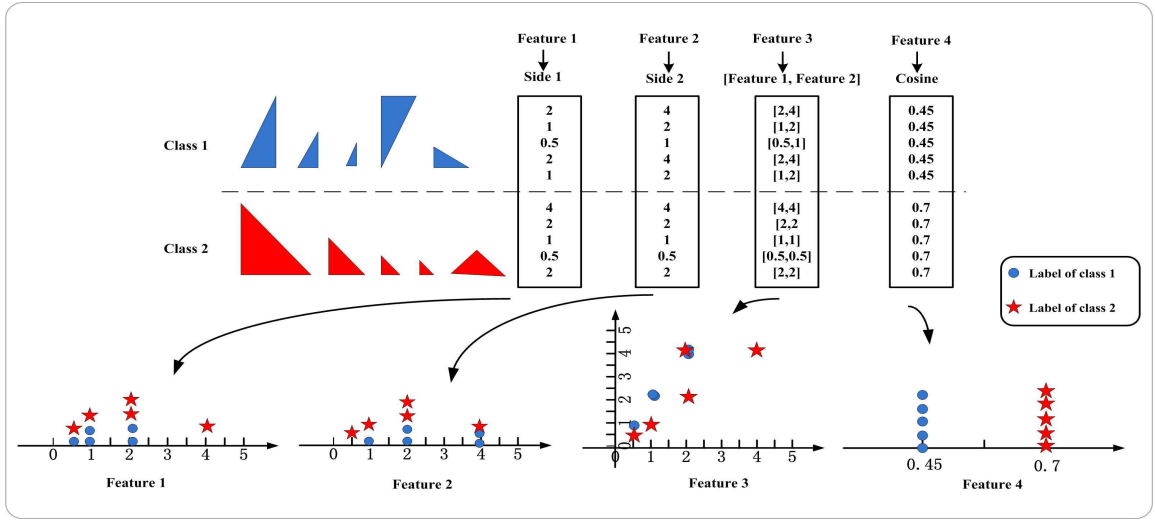}
\caption{The illustration of single feature, concatenating feature and nonlinear combined feature}
\label{corticalarchitecturefig}
\end{figure*}
Generally, many state-of-the-art methods only use a single type of feature, global or local, to represent the corresponding video, which causes difficulty in resisting various video attacks. For example, Fig. 1(a) shows the original picture while Fig. 1(b) shows the copy after some geometric and photometric variations. Obviously, the color and brightness distribution are different between these two pictures. If we only use a global feature such as a color or brightness histogram to generate the video fingerprint, we will obtain inaccurate results. By contrast, if we consider local features such as Harris interest points together, we may detect the copy accurately. A combination of multiple feature-based algorithms has been presented in recent years because of the limitation of individual features. Song et al. [18] proposed a video hashing learning method by weighted multiple feature fusion.
Jiang et al. [19] proposed a video copy detection system based on temporal pyramid matching based on two layers. Li et al. [20, 21] used SIFT, SURF, DCT coefficients, and weighted ASF to detect video copies, and made the final decision by fusing the detection results from different detectors. Although these methods combine multiple features to a certain extent, they do not fully use the assistance and consensus among features, which is important in multiple feature fusion.

Each type of feature reflects specific information on the original video, and can be taken as one view of the original video. According to the multi-viewing theory [36], the mutual information of different views (features) can improve the performance of the task. Therefore, a fusion of different features is beneficial to improve the performance of video fingerprinting. Intuitively, concatenating different feature vectors one by one is a directed form of feature fusion; moreover, combining multiple features with different weights is also a solution. However, the concatenated or weighted vector representation of features weakens the power of the propagation among the multiple features and even ignores their relations to a certain extent. For example, Fig. 2 illustrates a simple scenario of this issue. In this figure, two classes of right triangles exist and are similar with each other in each class (we can call them near-duplicates) because only affine transformations (scaling and rotation) are performed. We take the two sides of each right triangle as Feature 1 and Feature 2, respectively. Then, the concatenating feature [Feature 1, Feature 2] is labeled as Feature 3. We use a nonlinear combination cosine as the fourth feature which is ${\rm{cosine = (}}Feature{\rm{ 1)/(}}\sqrt {{{(Feature{\kern 1pt} {\kern 1pt} {\kern 1pt} 1)}^2} + {{(Feature{\kern 1pt} {\kern 1pt} {\kern 1pt} 2)}^2}} {\rm{)}}$. Obviously, the fourth feature cosine shows the best performance of classification, and proves that correlation and consensus that lead to performance improvement among different features exists.

Moreover, the concatenated or weighted feature vectors are difficult to handle with different scales. When imaging a scenario of people similarity evaluation, we use two different features, namely,\emph{ height} and \emph{weight} of people. If we concatenate these two features to a new feature [\emph{height, weight}], we may not evaluate the similarity of people correctly. For example, given two people with feature vector of [175 cm, 65 kg] and [160 cm, 80 kg], the first person is 15 cm taller but 15 kg lighter than the second person. Thus, evaluating these two people similar or not is difficult. Therefore, we should explore a new strategy to fuse multiple features, which can capture the inherent characteristic of the original data and the assistance between different features.

In addition, the environment of a network is increasingly complex, and combined modifications applied into videos are increasingly common. One of the challenges in NDVD is the robustness under the combined modifications. Generally, when traditional methods such as single feature-based and concatenation-based methods suffer some combined modifications, the performances are not as good as the performance under single-mode modification. In fact, we can model each feature as Eq. (1). Assume $V$ is the principle content of a video, then
\begin{equation}
V = f^{(i)}  + e^{(i)}
\end{equation}
where $f^{(i)}$ and $e^{(i)}$ are the $i_{th}$ feature of the video and its error, respectively. When the video suffers different modifications, the error is big or small, which demonstrates bad or good robustness. As is said, each feature is a view of the original data. According to the boundary of multi-view analysis [37], the connection between the consensus of two hypotheses on two views respectively
and their error rates is follows:
\begin{equation}
P(f^{(i)}  \ne f^{(j)} ) \ge \max \{ P_{err} (f^{(i)} ),P_{err} (f^{(j)} )\}
\end{equation}
The probability of a disagreement of two features upper bounds the error rate of either feature from the inequality. Thus, by exploring the
maximized agreement and consensus of different features, the error rate of each feature will be minimized. Therefore, mining the assistance and consensus among multiple features and making full use of them are necessary for NDVD.

To address the aforementioned issues, we propose a new notation called comprehensive feature to represent video content in this study. The comprehensive feature is not an original video feature, but a comprehensive, intrinsic, and transformed feature that can capture the assistance and consensus among multiple features. Compared with other types of features, the comprehensive feature has two advantages (detailed analysis is in the section 2.2.4): (1) It consists of the principal components and intrinsic
characteristics of the original video content, which can maximize the consensus of different features; (2) It is a compact and comprehensive representation with eliminating noises and uselessness information among different features. In this study, we propose a general comprehensive feature-mining scheme based on a tensor model. Tensor is the natural generalization of vector and matrix, and the algebra of higher-order tensors can consider the contextual different features. Furthermore, the tensor representation and decomposition can express intra-feature context correlation intuitively and propagate corresponding inter-feature context conveniently as well [22], because the tensor decomposition is a mixed staggered sampling, i.e., alternating sampling in the different components occurs rather than sampling from the components one by one. Therefore, the tensor model is one of the best tools for comprehensive feature generation. In summary, the main contributions of this study are the following:


(1) We present a comprehensive feature-based scheme to capture the assistance and consensus among different features using tensor model, and it is feasible to fusing any given features using the proposed scheme. In this scheme, the intrinsic and latent characteristics of multiple features are expressed completely through the comprehensive feature. We also show what we mined in the comprehensive feature, and give a theoretical analysis about its robustness.

(2) We propose an auxiliary matching strategy based on the tensor model. In this strategy, the core tensor is used to narrow the search range, and then the existing matching algorithm can be implemented in the smaller obtained dataset to find a match. This matching strategy can accelerate fingerprint matching to a certain extent, especially for a large-scale video fingerprint database.

Security and binarization are also considered in the video fingerprinting system. Randomization strategy [23, 24, 29] via secret keys is popularly used in the fingerprinting system to enhance the security, such as randomly selected image blocks [23] and overlapping sub-video cubes [23, 24]. The final fingerprint sequence can also be quantified to binary values via secret keys for binary quantization, and the video fingerprinting is also called video hashing [13], especially after the quantization such as in [6, 11, 18]. These common strategies can be definitely applied in the proposed method; for example, we can randomly select sub-video cubes of the original videos in the tensor decomposition, and quantify the final fingerprint vector into binary values via a key. These strategies and their analysis have been described in detail in the existing methods, so we do not discuss these issues in the present study.

\section{Proposed Method }
In this section, we present a roubust video fingerprinting scheme based on comprehensive feature using tensor model, and then we show an example to prove its performance. To make the scheme more understandable, we first list certain notations regarding the tensor [25].

\subsection{Related Notations and Technologies}
\textbf{Tensor}. A tensor  is a multidimensional array. Formally, an $N_{th}$-order tensor is an element of the tensor product of $N$ vector spaces, each with its own coordinate system. A vector and matrix are the first- and second-order tensors, respectively. Tensors of order three or higher are called higher-order tensors.

\textbf{Subarrays}.  Similar to matrices whose subarrays are rows and columns, subarrays are commonly used in tensor analysis. Fiber and slice are two subarrays of tensor. A fiber is defined by fixing every index but one. Slices are two-dimensional sections of a tensor, defined by fixing all but two indices.

\textbf{Matricization}. Matricization is a process of transforming an $N_{th}$-order tensor into a matrix. The mode-$n$ matricization of a tensor ${\bf{\chi }} \in {\Re ^{{I_1} \times {I_2} \times  \cdots  \times {I_N}}}$ is denoted by ${{\bf{X}}_{(n)}}$, and arranges the mode-$n$ fibers to be the columns of the resulting matrix.

\textbf{$n$-Mode Product}. The $n$-mode product can also be called tensor multiplication. The $n$-mode product of a tensor ${\bf{\chi }} \in {\Re ^{{I_1} \times {I_2} \times  \cdots  \times {I_N}}}$ with a matrix ${\bf{U}} \in {\Re ^{J \times {I_n}}}$ is denoted by ${\bf{\chi }} \times {}_n{\bf{U}}$ with size ${I_1} \times {I_2} \times  \cdots  \times {I_{n - 1}} \times J \times {I_{n + 1}} \times  \cdots  \times {I_N}$. That is
\begin{equation}
{({\bf{\chi }} \times {}_n{\bf{U}})_{{i_1} \cdots {i_{n - 1}}j{i_{n + 1}} \cdots {i_N}}} = \sum\limits_{{i_n} = 1}^{{I_n}} {{x_{{i_1}{i_2} \cdots {i_N}}}{u_{j{i_n}}}}
\end{equation}
where $x$ and $u$ are the elements of tensor ${\bf{\chi }}$ and matrix ${\bf{U}}$, respectively. Each mode-$n$ fiber is multiplied by matrix ${\bf{U}}$; thus, the idea can also be expressed as follows:
\begin{equation}
\begin{array}{l}
{\bf{\eta }} = {\bf{\chi }} \times {}_n{\bf{U}} \Leftrightarrow {{\bf{Y}}_{(n)}} = {\bf{U}}{{\bf{X}}_{(n)}}
\end{array}
\end{equation}
where ${{\bf{X}}_{(n)}}$ and ${{\bf{Y}}_{(n)}}$ are the mode-$n$ matricization of tensor ${\bf{\chi }}$ and ${\bf{\eta }}$, respectively.

High-order tensors can be approximated by the sum of low-rank tensors (the definition of tensor rank is described in [25]). The CANDECOMP/PARAFAC (CP) and Tucker model decomposition are two popular methods for approximation, which factorize a tensor into a sum of component low-rank tensors.

\textbf{CP model.} The CP model factorizes a tensor into a sum of component rank-one tensors. Given a third-order tensor ${\bf{\chi }} \in {\Re ^{I_1 \times I_2 \times I_3}}$, it can be concisely expressed as
\begin{equation}
{\bf{\chi }} \approx \left[\kern-0.15em\left[ {{\bf{\lambda }}; {\kern 1pt} {\kern 1pt} {{\bf{A}}^{(1)}},{\kern 1pt} {\kern 1pt} {\kern 1pt} {{\bf{A}}^{(2)}},{\kern 1pt} {\kern 1pt} {\kern 1pt} {{\bf{A}}^{(3)}}{\kern 1pt} }
 \right]\kern-0.15em\right] = \sum\limits_{r = 1}^R {{\lambda _r}{\bf{a}}_r^{(1)} \circ {\bf{a}}_r^{(2)} \circ {\bf{a}}_r^{(3)}}
\end{equation}
where $R$ is a positive integer and ${{\bf{a}}_r^{(n)}} \in {\Re ^{I_n}}$, and $\lambda _r$ is constant for $r = 1, \cdots ,R$. ${{\bf{A}}^{(n)}} = [{{{\bf{a}}}_1^{(n)}},{{{\bf{a}}}_2^{(n)}}, \cdots ,{{{\bf{a}}}_R^{(n)}}]$ with $1 \le n \le 3$, and are factor matrices of the tensor. The symbol ``$\circ$" represents the vector outer product.

\textbf{Tucker model.} The Tucker model decomposes a tensor into a core tensor multiplied by matrices along each mode. For example, a third-order tensor ${\bf{\chi }}$, can be decomposed as follows:
\begin{equation}
\begin{array}{l}
{\bf{\chi }} \approx \left[\kern-0.15em\left[ {{\bf{\kappa }};{{\bf{A}}^{(1)}},{{\bf{A}}^{(2)}},{{\bf{A}}^{(3)}}}
 \right]\kern-0.15em\right] = {\bf{\kappa }}{ \times _1}{{\bf{A}}^{(1)}}{ \times _2}{{\bf{A}}^{(2)}}{ \times _3}{{\bf{A}}^{(3)}}\\
 = \sum\limits_{{j_1} = 1}^{{J_1}} {\sum\limits_{{j_2} = 1}^{{J_2}} {\sum\limits_{{j_3} = 1}^{{J_3}} {{k_{{j_1}{j_2}{j_3}}}{\bf{a}}_{{j_1}}^{(1)} \circ {\bf{a}}_{{j_2}}^{(2)}}  \circ {\bf{a}}_{{j_3}}^{(3)}} } \\
where{\kern 1pt} {\kern 1pt} {\kern 1pt} {\kern 1pt} {\kern 1pt} {{\bf{A}}^{(n)}} = [{\bf{a}}_1^{(n)},{\bf{a}}_2^{(n)}, \cdots ,{\bf{a}}_{{J_n}}^{(n)}] \in {\Re ^{{I_n} \times {J_n}}},{\kern 1pt} \\
{\kern 1pt} {\kern 1pt} {\kern 1pt} {\kern 1pt} {\kern 1pt} {\kern 1pt} {\kern 1pt} {\kern 1pt} {\kern 1pt} {\kern 1pt} {\kern 1pt} {\kern 1pt} {\kern 1pt} {\kern 1pt} {\kern 1pt} {\kern 1pt} {\kern 1pt} {\kern 1pt} {\kern 1pt} {\kern 1pt} {\kern 1pt} {\kern 1pt} {\kern 1pt} {\kern 1pt} 1 \le n \le 3
\end{array}
\end{equation}
where $I_n$ and $J_n$ are positive integers, and ${\bf{\kappa }}$ is the core tensor. ${\bf{A}}_{(n)}$ is the factor matrix.

Tensor has been applied to many fields, such as computer vision and signal processing. In this study, the video is considered a third-order tensor in the tensor space constructed by different features.

\subsection{Comprehensive Feature Mining-based Robust Video Fingerprinting}
 In the proposed scheme, we first use different features to generate a video tensor. Then, the comprehensive feature is mined to generate the video fingerprint using Tucker model. In addition, a matching strategy is presented by the core tensor to accelerate fingerprint matching.

\subsubsection{Video Tensor Construction}
We let $M$ be the number of features, and ${\bf{x}}^m  \in \Re ^{d_m  \times 1}$ the $m_{th}$ feature, where $m = 1,2, \cdots ,M$ and $d_m$ is the dimensionality of the $m_{th}$ video feature. The video is considered a third-order tensor
constructed by the $M$ features showed in Fig. 3. In the video tensor, the frontal slice is the multiple feature values, whereas mode-3 is the time sequence.
The length of each feature vector and the total number of features determine the size of the frontal slice. One feasible strategy is to take the mean length of the entire feature vector as the standard length, and then concatenate all feature vectors column by column in the front slice with standard length for the column. Then, we use tensor decomposition to generate the comprehensive feature.
\begin{figure}[htbp] \centering
\includegraphics [width=2.5in]{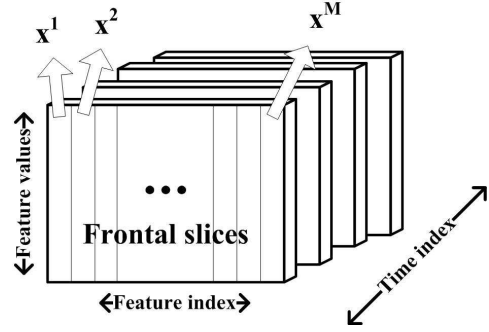}
\caption{The diagram of the video tensor}
\label{corticalarchitecturefig}
\end{figure}


\subsubsection{Tensor Model Selection}
 The CP and Tucker models are two main tensor decomposition models. We use the Tucker model in the proposed framework. The third-order Tucker model represents the data spanning the three modes by the vectors given by columns ${\bf{A}}^{(1)}$, ${\bf{A}}^{(2)}$ and ${\bf{A}}^{(3)}$ as shown in Eq. (6). As a result, the Tucker model encompasses all possible linear interactions between vectors pertaining to the various modes of the data [28]. The CP model is a special case of the Tucker model where the size of each mode is the same, i.e. $J_1 = J_2 = J_3$ in Eq. (6). In this study, we used the Tucker model instead of the CP model. The first reason is that the Tucker model is more flexible and scalable during decomposition, which allows users to select different numbers of factors along each mode. The CP decomposition can only provide a particular tensor with a particular number of components when the component loadings are highly correlated in all the modes, i.e., the results of CP decomposition are mathematical artifact that is not physically meaningful. There are strong effects among the various components decomposed by CP model, which make the CP model unstable, slow in convergence, and difficult to interpret [30]. More importantly, a core tensor, which considers all possible linear interactions between the components of each mode, can be obtained by the Tucker model, and is stable [25, 30] to a certain extent. The core tensor is used for matching in the proposed framework.

Choosing the number of components specified for each mode (i.e., the values of $J_1$, $J_2$ and $J_3$ in Eq. (6))  is a particular challenge in the Tucker model. In the present study, we use a Bayesian approach called automatic relevance determination (ARD) that described in [28] to determine the number of components in each mode. ARD alternately optimizes the
parameters to discover which components are relevant, and the parameters are modeled as either Gaussian prior or Laplace prior. The Gaussian prior $P_G$ and Laplace prior $P_L$ on the parameter ${\bf{\theta }}_d$ are following.
\begin{equation}
P_G ({\bf{\theta }}_d |\alpha _d ) = \prod\limits_j {(\frac{{\alpha _d }}{{2\pi }})^{1/2} } \exp ( - \frac{{\alpha _d }}{2}\theta _{j,d}^2 )
\end{equation}
\begin{equation}
{P_L}({\bf{\theta}} _d|{\alpha _d}) = \prod\limits_j {\frac{{{\alpha _d}}}{2}} \exp ( - {\alpha _d}|\theta _{j,d}^{}|_1)
\end{equation}

We will describe how we use ARD to determine the values of $J_1$, $J_2$ and $J_3$ in the proposed method.

We let ${\bf{\chi }}$ be the third-order video tensor, then, for the Tucker decomposition,
\begin{equation}
{\bf{\chi }} \approx {\bf{\Gamma }} = {\bf{\kappa }} \times _1 {\bf{A}}^{(1)}  \times _2 {\bf{A}}^{(2)}  \times _3 {\bf{A}}^{(3)}
\end{equation}

The model can also be written as
\begin{equation}
\begin{array}{*{20}c}
   {\chi _{i_1 ,i_2 ,i_3 }  \approx \Gamma _{i_1 ,i_2 ,i_3 } }  \\
   { = \sum\limits_{j_1 ,j_2 ,j_3 } {\kappa _{j_1 ,j_2 ,j_3 } {\bf{A}}_{i_1 ,j_1 }^{(1)} {\bf{A}}_{i_2 ,j_2 }^{(2)} {\bf{A}}_{i_3 ,j_3 }^{(3)} } }  \\
\end{array}
\end{equation}
where ${\bf{\kappa }} \in {\bf{R}}^{J_1  \times J_2  \times J_3 }$ and ${\bf{A}}^{(n)}  \in {\bf{R}}^{I_n  \times J_n}$. If we denote this Tucker model as $T(J_1 ,J_2 ,J_3 )$, our goal is to find the optimal values of $J_1$, $J_2$ and $J_3$.

According to Eq. (9), the tensor $\bf{\chi }$ can also be rewritten as
\begin{equation}
{\bf{\chi }} ={\bf{\Gamma }}  + {\bf{\xi }}
\end{equation}
where $\bf{\xi }$ is an error parameter, and its distribution can be taken as an independent identically distribution (i.i.d.) with Gaussian noise. i.e.,
\begin{equation}
\begin{array}{*{20}c}
   {P(\xi ) = P(\chi |{\bf{\Gamma }},\sigma )}  \\
   { = \prod\limits_{i_1 ,i_2 , i_3 } {\frac{1}{{\sqrt {2\pi \sigma ^2 } }}} \exp ( - \frac{({\chi _{i_1 ,i_2 ,i_3 }  - \Gamma _{i_1 ,i_2 , i_3 } })^2}{{2\sigma ^2 }})}  \\
\end{array}
\end{equation}

Therefore, we can explore optimal $J_1$, $J_2$ and $J_3$ by minimizing the least squares objective $||{\bf{\chi}}-{\bf{\Gamma}}||_F^2$. In the Bayesian framework, it corresponds to minimizing the negative log-likelihood.

According to Eq. (7) and (8), we obtain ${P_G}({{\bf{A}}^{(n)}}|{\alpha ^{(n)}})$, $P_L ({\bf{A}}^{(n)} |{\bf{\alpha }}^{(n)})$, ${P_G}({\bf{\kappa }}|{{\bf{\alpha }}^\kappa })$ and ${P_L}({\bf{\kappa }}|{{\bf{\alpha }}^\kappa })$ by Gaussian or Laplace prior, respectively, where $1 \le n \le 3$. The parameters ${{\bf{\alpha }}^{(n)}}$ and ${{\bf{\alpha }}^\kappa }$ are also given as uniform priors for simplification in ARD. As a result, the posterior can be
written as
\begin{equation}
\begin{array}{l}
L = P({\bf{\kappa }},{{\bf{A}}^{(1)}},{{\bf{A}}^{(2)}},{{\bf{A}}^{(3)}}|{\bf{\chi }},\sigma ,{{\bf{\alpha }}^\kappa },{{\bf{\alpha }}^{(1)}}, {{\bf{\alpha }}^{(2)}}, {{\bf{\alpha }}^{(3)}})\propto \\
 P({\bf{\chi }}|{\bf{\Gamma }} ,\sigma^2 )P({\bf{\kappa }}|{{\bf{\alpha }}^\kappa })P({{\bf{A}}^{(1)}}|{{\bf{\alpha }}^{(1)}})P({{\bf{A}}^{(2)}}|{{\bf{\alpha }}^{(2)}})P({{\bf{A}}^{(3)}}|{{\bf{\alpha }}^{(3)}})
\end{array}
\end{equation}

Subsequently, the negative log likelihood ($- {\mathop{\rm Log}\nolimits} (L)$) using Gaussian and Laplace priors are proportional to the Eqs. (14) and (15), respectively.
\begin{equation}
\begin{array}{l}
 C + \frac{1}{{2{\sigma ^2}}}||{\bf{\chi }} -  \Gamma  |{|_F^2} + 0.5*\sum\limits_n {\sum\limits_d {\alpha _d^{(n)}||{\bf{A}}_d^{(n)}||_F^2} }+ {\alpha ^\kappa }||{\bf{\kappa }}||_F^2 \\
 + 0.5*{I_1}{I_2}{I_3}\log {\sigma ^2} - 0.5*\sum\limits_n {\sum\limits_d {{I_n}\log \alpha _d^{(n)}} } \\
 - 0.5*{J_1}{J_2}{J_3}\log {\alpha ^\kappa }
\end{array}
\end{equation}
\begin{equation}
\begin{array}{l}
C + \frac{1}{{2{\sigma ^2}}}||{\bf{\chi }} - \Gamma |{|_F^2} + \sum\limits_n {\sum\limits_d {\alpha _d^{(n)}||{\bf{A}}_d^{(n)}||_1^{}} }+ {\alpha ^\kappa }||{\bf{\kappa }}|{|_1} \\
+0.5*{I_1}{I_2}{I_3}\log {\sigma ^2} - \sum\limits_n {\sum\limits_d {{I_n}\log \alpha _d^{(n)}} } \\
 - {J_1}{J_2}{J_3}\log {\alpha ^\kappa }
\end{array}
\end{equation}
where $C$, $|| \bullet |{|_F}$ and $|| \bullet |{|_1}$ are a constant value, F-norm and 1-norm, respectively.

We equate the derivatives of Eq. (14) with respect to $\sigma ^2$, $\alpha _d^{(n)}$ and $\alpha ^\kappa$ to zero, respectively. Then we obtained  the following parameters by Gaussian prior
\begin{equation}
{\sigma ^2} = \frac{{||{\bf{\chi }} - \Gamma ||_F^2}}{{{I_1}{I_2}{I_3}}},{\kern 1pt} {\kern 1pt} {\kern 1pt} {\kern 1pt} \alpha _d^{(n)} = \frac{{{I_n}}}{{||{\bf{A}}_d^{(n)}||_F^2}},{\kern 1pt} {\kern 1pt} {\kern 1pt} {\kern 1pt} {\alpha ^\kappa } = \frac{{{J_1}{J_2}{J_3}}}{{||\kappa ||_F^2}}
\end{equation}

The parameters by Laplace prior can also be obtained in the same way by equating the derivatives of Eq. (15) to zero, which are as follows.
 \begin{equation}
{\sigma ^2} = \frac{{||{\bf{\chi }} - \Gamma ||_F^2}}{{{I_1}{I_2}{I_3}}},{\kern 1pt} {\kern 1pt} {\kern 1pt} {\kern 1pt} \alpha _d^{(n)} = \frac{{{I_n}}}{{||{\bf{A}}_d^{(n)}||_1}},{\kern 1pt} {\kern 1pt} {\kern 1pt} {\kern 1pt} {\alpha ^\kappa } = \frac{{{J_1}{J_2}{J_3}}}{{||\kappa ||_1}}
\end{equation}

Generally, the first row expressions of Eqs. (14) and (15) can be considered as $l_2$-regularized and $l_1$-regularized problems respectively, while the second and third rows are the normalization constants in the likelihood terms. The $l_2$-regularized and $l_1$-regularized problems are equivalent to the regular ridge regression and sparse regression problems that can solved by existing algorithms, respectively.

In Eqs. (16) and (17), ${\sigma ^2}$ can be learned from data or estimated by a signal-to-noise rate (SNR)-based method [28]. Given the initialization of the parameters and ${{\bf{A}}^{(n)}}$, The optimal values of $J_1$, $J_2$ and $J_3$ are obtained by alternately estimating $\alpha _d^{(n)}$, ${\kern 1pt} {\alpha ^\kappa }$, $\kappa$ and ${{\bf{A}}^{(n)}}$ based on Eqs. (16-17) and the solutions of two regularized problems until convergence. Which prior assumption (Gaussian or Laplace) is used to update the parameters is corresponding to setting the parameters of the priors such that
they match the posteriors distribution [38].

\subsubsection{Comprehensive Feature Mining}
After the video tensor decomposition by Tucker model, three low-rank tensor matrices ${\bf{A}}^{(n)} (n = 1,2,3)$ are obtained, which fused the different video features. To make the final fingerprint vector more compact, we average the absolute values of elements in component matrix ${{\bf{A}}^{(n)}}$ by row to obtain a component vector, and then concatenate the three component vectors to obtain the comprehensive feature $\bf{y}$ which is considered as the final video fingerprint vector.

The step by step description of the proposed framework is provided in Algorithm 1.
\begin{algorithm}
\caption{Comprehensive feature based video fingerprinting} \label{alg1}
\algsetup{indent=2em}
\algsetup{linenosize=\tiny}
\footnotesize
\begin{algorithmic}
\STATE \textbf{Input:} Query video $V_{q}$.
\STATE \textbf{Output:} A video fingerprint vector $\bf{y}$.
\STATE \textbf{(1) Video tensor construction:} Given multiple features such as global, local and temporal features, a video tensor ${\bf{\chi }}$ is constructed with size $I_1 \times I_2 \times I_3$.
\STATE \textbf{(2) Tensor decomposition:} The video tensor is decomposed by the Tucker model with optimized number of components selected based on the ARD-based algorithm, and we obtain three modes of components, which are ${{\bf{A}}^{(1)}} = {(a_{i_1j_1}^{(1)})_{I_1 \times J_1}}$, ${{\bf{A}}^{(2)}} = {(a_{i_2j_2}^{(2)})_{I_2 \times J_2}}$ and ${{\bf{A}}^{(3)}} = {(a_{i_3j_3}^{(3)})_{I_3 \times J_3}}$, respectively.
\STATE \textbf{(3) Fingerprint calculation:} We average the absolute values of elements in component matrix ${{\bf{A}}^{(n)}}$ ($n=1,2,3$) by row to obtain a component vector, and then concatenate the three component vectors to obtain the comprehensive feature vector $\bf{y}$ which is taken as the video fingerprint vector.

${\bf{y}} = [\frac{1}{J_1}\sum\limits_{j_1 = 1}^{J_1} {|a_{i_1j_1}^{(1)}|{\kern 1pt} {\kern 1pt} {\kern 1pt} {\kern 1pt} ;\frac{1}{J_2}\sum\limits_{j_2 = 1}^{J_2} {|a_{i_2j_2}^{(2)}|{\kern 1pt} {\kern 1pt} {\kern 1pt} {\kern 1pt} ;} \frac{1}{J_3}\sum\limits_{j_3 = 1}^{J_3} {|a_{i_3j_3}^{(3)}|} } ]$
\end{algorithmic}
\end{algorithm}

\subsubsection {Analysis of Comprehensive Feature}
What we mined in the comprehensive feature is important in evaluating the performance of the proposed method. As mentioned, the Tucker model is used in comprehensive feature mining. We first review the process of Tucker decomposition.

Given a video tensor ${\bf{\chi }} \in \Re ^{I_1  \times I_2 \times I_3 }$, it can be decomposed as
\begin{equation}
\begin{array}{l}
 \chi  \approx \kappa  \times _1 {\bf{A}}^{(1)}  \times _2 {\bf{A}}^{(2)} \times _N {\bf{A}}^{(3)}  \\
  = \left[\kern-0.15em\left[ {\kappa ;{\bf{A}}^{(1)} ,{\bf{A}}^{(2)}, {\bf{A}}^{(3)} }
 \right]\kern-0.15em\right] \\
 \end{array}
\end{equation}
where each $ {\bf{A}}^{(n)}$ ($1 \le n \le 3$) is an orthogonal matrix, and $\kappa$ is the core tensor. For distinct modes
in a series of multiplications, the order of the multiplication is irrelevant [25]. Therefore, the core tensor $\kappa$ is
\begin{equation}
\kappa  \approx {\bf{\chi }} \times _1 {\bf{A}}^{(1)T}  \times _2 {\bf{A}}^{(2)T} \times _N {\bf{A}}^{(3)T}
\end{equation}
where ${\bf{A}}^{(n)T}$ is the transposition of ${\bf{A}}^{(n)}$. We can use the following optimization problem to obtain $\kappa$ and $ {\bf{A}}^{(n)}$.
\begin{equation}
\begin{array}{l}
 \mathop {\min }\limits_{\kappa ,{\bf{A}}^{(1)} {\bf{A}}^{(2)} {\bf{A}}^{(3)} } ||{\bf{\chi }} - \left[\kern-0.15em\left[ {\kappa ;{\bf{A}}^{(1)} ,{\bf{A}}^{(2)}, {\bf{A}}^{(3)} }
 \right]\kern-0.15em\right]||^2 \\
 s.t.{\kern 1pt} {\kern 1pt} {\kern 1pt} \kappa  \in \Re ^{J_1  \times J_2   \times J_3 } ,{\kern 1pt} {\kern 1pt} {\kern 1pt} {\kern 1pt}  \\
 {\kern 1pt} {\kern 1pt} {\kern 1pt} {\kern 1pt} {\kern 1pt} {\kern 1pt} {\kern 1pt} {\kern 1pt} {\kern 1pt} {\kern 1pt} {\kern 1pt} {\kern 1pt} {\kern 1pt} {\kern 1pt} {\bf{A}}^{(n)}  \in \Re ^{I_n  \times J_n } {\kern 1pt} and{\kern 1pt} {\kern 1pt} column-wise{\kern 1pt} {\kern 1pt} {\kern 1pt} orthogonal \\
 \end{array}
\end{equation}

Consequently, the objective function of Eq. (20) can be rewritten as Eq. (21).
\begin{equation}
\begin{array}{l}
 ||{\bf{\chi }} - \left[\kern-0.15em\left[ {\kappa ;{\bf{A}}^{(1)} ,{\bf{A}}^{(2)}, {\bf{A}}^{(3)} }
 \right]\kern-0.15em\right]||^2  \\
  = ||{\bf{\chi }}||^2  - 2 < {\bf{\chi }},\left[\kern-0.15em\left[ {\kappa ;{\bf{A}}^{(1)} ,{\bf{A}}^{(2)}, {\bf{A}}^{(3)} }
 \right]\kern-0.15em\right] >  \\
 \quad+ ||\left[\kern-0.15em\left[ {\kappa ;{\bf{A}}^{(1)} ,{\bf{A}}^{(2)}, {\bf{A}}^{(3)} }
 \right]\kern-0.15em\right]||^2  \\
  = ||{\bf{\chi }}||^2  - 2 < {\bf{\chi }} \times _1 {\bf{A}}^{(1)T}  \times _2 {\bf{A}}^{(2)T}  \times _3 {\bf{A}}^{(3)T} ,\kappa  >  + ||\kappa ||^2  \\
  = ||{\bf{\chi }}||^2  - 2 < \kappa ,\kappa  >  + ||\kappa ||^2  \\
  = ||{\bf{\chi }}||^2  - ||\kappa ||^2  \\
  = ||{\bf{\chi }}||^2  - ||{\bf{\chi }} \times _1 {\bf{A}}^{(1)T}  \times _2 {\bf{A}}^{(2)T}  \times _3 {\bf{A}}^{(3)T} ||^2  \\
 \end{array}
\end{equation}

Furthermore, the optimization problem Eq. (21) is equal to the following maximization problem  because $||\chi||^2$ is constant:
\begin{equation}
\begin{array}{l}
 \mathop {\max }\limits_{{\bf{A}}^{(n)} } ||{\bf{\chi }} \times _1 {\bf{A}}^{(1)T}  \times _2 {\bf{A}}^{(1)T} \times _3 {\bf{A}}^{(3)T} ||^2 \\
  \Leftrightarrow \mathop {\max }\limits_{{\bf{A}}^{(n)}} ||{\bf{A}}^{(n)T} {\bf{W}}|| \\
 s.t.{\kern 1pt} {\kern 1pt} {\kern 1pt} {\kern 1pt} {\kern 1pt} {\bf{W}} = {\bf{X}}_{(n)} ({\bf{A}}^{(3)}  \otimes {\bf{A}}^{(n+1)} \otimes {\bf{A}}^{(n-1)} \otimes {\bf{A}}^{(1)} ) \\
 \quad 1 \le n \le 3
 \end{array}
\end{equation}
where ``$\otimes$" is Kronecker product, and setting $\bf{A^{(4)}}=\bf{A^{(0)}}=\bf{I}$. This maximization can be solved using the ALS algorithm by setting ${\bf{A}}^{(n)}$ as the $J_n$ leading
left singular vectors of $\bf{W}$, and it will converge to a solution [35].

The final low-rank tensor matrix ${\bf{A}}^{(n)}$ is computed from the left singular vectors of $\bf{W}$ that consists of the original video tensor $\bf{\chi}$. According to the property of Singular Value Decomposition (SVD), the left matrix obtained from the SVD contains the intrinsic characteristics of the original matrix in the column space which consists of different features.

From another perspective, each column of ${\bf{W}}$ is an approximation of the feature vector. If each feature vector is taken as a point in original space, the principal components that represent the main information among the point distribution can be computed by Principal Component Analysis (PCA), and they are computed from the left singular matrix of ${\bf{W}}$ according to the relation between PCA and SVD. Therefore, the comprehensive feature mined from ${\bf{A}}^{(n)}$ can be considered the intrinsic and principal information of original data, where the noise and uselessness among different features have been eliminated. Generally, the different features should reach a consensus in representing the video content [36]. The intrinsic and principal information of the original video exploited during the comprehensive feature mining are only the consensus and assistance of different features, which lead to improved performance in representing the video content.

Considering the preceding analysis, we can conclude that the information exploited during the comprehensive feature mining is composed of the principal components and intrinsic characteristics of the original video tensor. However, the comprehensive feature mining is different from SVD or PCA because the extraction of the principal components and intrinsic characteristics in comprehensive feature mining is iterative until converge while it is one-off in SVD or PCA. The iterative process and alternative optimization capture more intrinsic information and consensus, which is the reason we use the tensor model during comprehensive feature mining.

\subsubsection {Robustness of Comprehensive Feature}
Robustness is the most important issue in video fingerprinting system. In this section, we present a qualitative analysis of the proposed framework. The experimental result in the next section proves the robustness of the comprehensive feature.

The video fingerprint vector of the proposed method is the comprehensive feature that consists of three modes of video tensors. Therefore, without loss of generality, the mode-3 factor matrix ${\bf {A}}^{(3)}$ is taken as an example to analyze the robustness. Given that $\chi$ is the tensor of a video, the approximation of this tensor through the Tucker model is
\begin{equation}
\begin{array}{*{20}{l}}
{{\bf{\chi }} \approx {\bf{\kappa }}{ \times _1}{{\bf{A}}^{(1)}}{ \times _2}{{\bf{A}}^{(2)}}{ \times _3}{{\bf{A}}^{(3)}}}\\
{ = {\bf{\gamma }}{ \times _3}{{\bf{A}}^{(3)}}}
\end{array}
\end{equation}
where ${\bf{\gamma }} = {\bf{\kappa }} \times {}_1{{\bf{A}}^{(1)}} \times {}_2{{\bf{A}}^{(2)}}$. According to the tensor multiplication and matricization, Eq. (23) can be written by matrices:
\begin{equation}
{\bf{X}} \approx {{\bf{A}}^{(3)}{\bf{P}}}
\end{equation}
where ${{\bf{X}}}$ and ${{\bf{P}}}$ are the matricizations of ${\bf{\chi }}$ and ${\bf{\gamma }}$ in mode-3, respectively. ${{\bf{X}}}$ is the multiple feature matrix in the proposed method. Furthermore, we rewrite Eq. (24) as
\begin{equation}
{{\bf{P}}^T}{{\bf{A}}^{(3)T}} \approx {{{\bf{X}}^T}}
\end{equation}

Obviously, we can  approximately consider Eq. (25) as a linear equation Eq. (26), where ${\bf{P}}^T$, ${\bf {A}}^{(3)T}$ and ${{\bf{X}}^T}$ are coefficient matrix, variable and constant term, respectively.
 \begin{equation}
{{\bf{P}}^T}{{\bf{A}}^{(3)T}} = {{{\bf{X}}^T}}
\end{equation}

 Now, we can consider the robustness of comprehensive feature ${\bf{A}}^{(3)}$ as the stability of the solution in this linear equation. Therefore, if the solution of Eq. (26) is stable, ${\bf{A}}^{(3)}$ becomes robust.
The metric that measures  the solution stability of a linear equation is the condition number of coefficient matrix. Smaller condition numbers mean stronger stability of the solution. Then, the condition number of the coefficient matrix in Eq. (26) is
\begin{equation}
{{\mathop{\rm condition}\nolimits}}_2({{\bf{P}}^T}) \approx {{\mathop{\rm condition}\nolimits}}_2({(\kappa  \times {}_1{\bf{A}}^{(1)} \times {}_2{\bf{A}}^{(2)})^T})
\end{equation}
where ${{\mathop{\rm condition}\nolimits}}_2( \bullet )$ is the 2-norm condition number of $\bullet$, and it is defined as ${{\mathop{\rm condition}\nolimits}}_2( \bullet ) = || \bullet |{|}_2||{ \bullet ^{ - 1}}|{|}_2$.
${\bf{A}}^{(1)}$ and ${\bf{A}}^{(2)}$ are the factor matrices after the Tucker decomposition, which are column-wise orthogonal [25]. According to the property of condition
(i.e., ${{\mathop{\rm condition}\nolimits} _2}({\bf{UM}}) = {{\mathop{\rm condition}\nolimits} _2}({\bf{M}})$, if $\bf{U}$ is an orthogonal matrix), thus, Eq. (27) is rewritten as
\begin{equation}
\begin{array}{l}
 {\mathop{\rm condition}\nolimits} _2 ((\kappa  \times {\bf{A}}^{(1)}  \times {\bf{A}}^{(2)} )^T ) \\
  = {\mathop{\rm condition}\nolimits} _2 (({\bf{A}}^{(2)} )^T  \times ({\bf{A}}^{(1)} )^T  \times \kappa ^T ) = {\mathop{\rm condition}\nolimits} _2 (\kappa ^T ) \\
 \end{array}
\end{equation}

 As mentioned, $\kappa^T$ is the transposition of the core tensor, and its matricization is a non-singular matrix. Generally, its condition is small. Fig. 4 shows the distribution of theire condition numbers. Most of the condition numbers are between 2 and 18, which are acceptable values for a well-conditioned problem.
Therefore, Eq. (26) is a well-conditioned equation, and has a stable solution when the coefficient matrix or constant term changed. In other words,  ${\bf {A}}^{(3)}$,
which is used to generate the final video fingerprint is robust.
\begin{figure}[htbp] \centering
\includegraphics [width=2.8in]{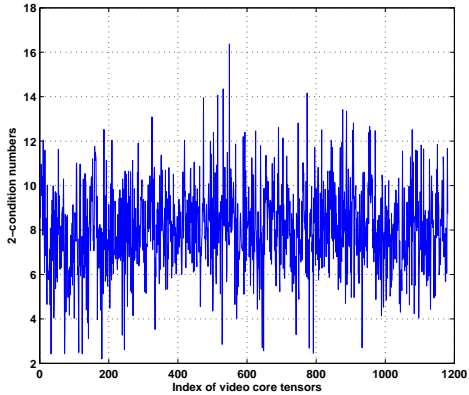}
\caption{condition numbers of the transposition of video core tensor}
\label{corticalarchitecturefig}
\end{figure}

\subsubsection{Matching Strategy}
Fingerprint vector matching is one of the important steps in a video fingerprinting system. In real application, the video website manages a video fingerprint database. When a user uploads a new video, the management first generates its fingerprint and then matches the fingerprints in the database. Many matching schemes exist in the literature, such as exhaustive matching, tree-based strategy and inverted files. In this study, we provide an auxiliary strategy for the existing matching schemes. The goal of this strategy is to narrow
the matching range in advance, and then the existing matching schemes can be executed in the obtained dataset with smaller sizes compared with the original.
The proposed matching strategy is based on the core tensor $ {\bf{\kappa }}$, which represents the level of interaction between the different components.

In this strategy, we use the sum of elements in the core tensor as a match tag, and the pre-matching is conducted in the database using the tag. Given a query video fingerprint with match tag ${T_q}$, the fingerprint, whose match tags are between ${T_q} - a*{T_q}$ and ${T_q} + a*{T_q}$, are obtained in the database after the pre-matching.
  Because the length of the match tag is one, the pre-matching is rapid. The parameter $a \in [0,1]$ is called the adjustment factor. We show how to select the adjustment factor in the next paragraph.

Given that $S$ is the fingerprint dataset obtained after the pre-matching, ${P_s}$ is the the rate of falling into $S$ for the visual-similar videos (which are the true near-duplicates we want to find), while ${P_{nd}}$ is the rate of not falling into $S$ for the visual-different videos (which are not the near duplicates). Intuitively, we could expect
that both ${P_s}$ and ${P_{nd}}$ are high. However, the larger $a$ causes a higher ${P_s}$ but lower ${P_{nd}}$. Conversely, a smaller $a$ leads
to a high ${P_{nd}}$ but lower ${P_s}$. Therefore, a tradeoff should be considered between ${P_s}$ and ${P_{nd}}$ by choosing the adjustment factor $a$. Given a video dataset, we can use
the curves of ${P_s}$ and ${P_{nd}}$ versus $a$ to choose the adjustment factor. Fig. 5 shows ${P_s}$ and ${P_{nd}}$ versus
the adjustment factor $a$ for a video database. The intersection of the two curves is a good choice for the adjustment factor. However, we cannot use that
point because we should first consider avoiding a mismatch in the pre-matching, i.e., the fingerprint vectors similar to the query should be involved in $S$ as much as
 possible. ${P_s}$ is not sufficiently high (less than 90\%) on the intersection, which leads to a mismatch. Therefore, we choose $a$ with giving priority to
 $P_S$. Taking Fig. 5 as an example, we choose $a = 0.3$, where the ${P_s}$ is almost 1, and ${P_{nd}}$ is more than 0.6. In other words, almost all of the
  fingerprint vectors of visual similar videos compared with the query are involved in $S$ after pre-matching, and the ones of visual-different videos are not
  involved in $S$ with a rate more than 0.6. Only less than 40\% of the total visually different video fingerprint vectors fall
  into $S$. Therefore, the size of the dataset after pre-matching is reduced. Further matching can be performed by any existing search technology.
  Generally, we assume that $N$ videos exist in the original database, and the complex of the matching is reduced to $O(K)$, where $K$ is the number of fingerprint vectors after the pre-matching, and
  $K < N$.
\begin{figure}[htbp] \centering
\includegraphics [width=2.8in]{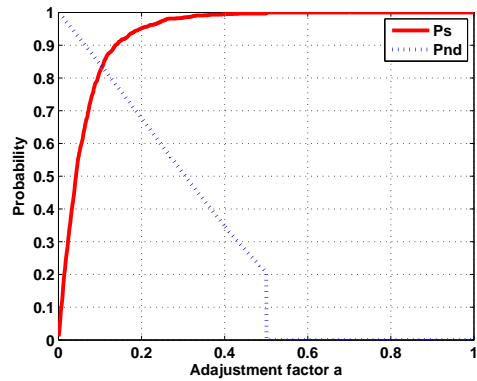}
\caption{The values of ${P_s}$ , ${P_{nd}}$ under different adjustment factors}
\label{corticalarchitecturefig}
\end{figure}

\subsubsection{Implementation of Proposed Scheme}
In this section, we show an implementation of the proposed scheme using global, local and temporal video features, i.e., $M=3$. These three types of features are
described as follows. However, fusing any type of features in the proposed scheme is feasible. In addition, we can also implement feature selection
strategies before inputting them to the system to eliminate some similar or conflicting features, but this issue is beyond the scope of the proposed framework.

 Histograms of different patterns, such as color, ordinal and block gradients, are common global features. We use normalized gray histograms
 as the global feature. Other global features can also be used in the proposed method.

Many descriptors of interest points can be taken as the local features. In this study, we use SURF descriptor as the local feature
that is invariant to scale, rotation and brightness variation [26]. Each SURF point in an image is associated with a 64-dimension descriptor, which focuses on the spatial distribution of gradient information within the interest point neighborhood.

In addition, we also compute the differences of histograms of pixels between adjacent frames, and then normalize them as a feature, which represent the variance
along time evolution. This feature can be considered as a kind of global feature, however, we list it separately in this study because it shows the temporal information of a video.

After extracting the multiple features of the video, we then take them as inputs for the proposed framework. The comprehensive feature is obtained for the final fingerprint vector. The experimental results are shown in Section 3.

\begin{figure*}[htbp] \centering
\includegraphics [width=5in]{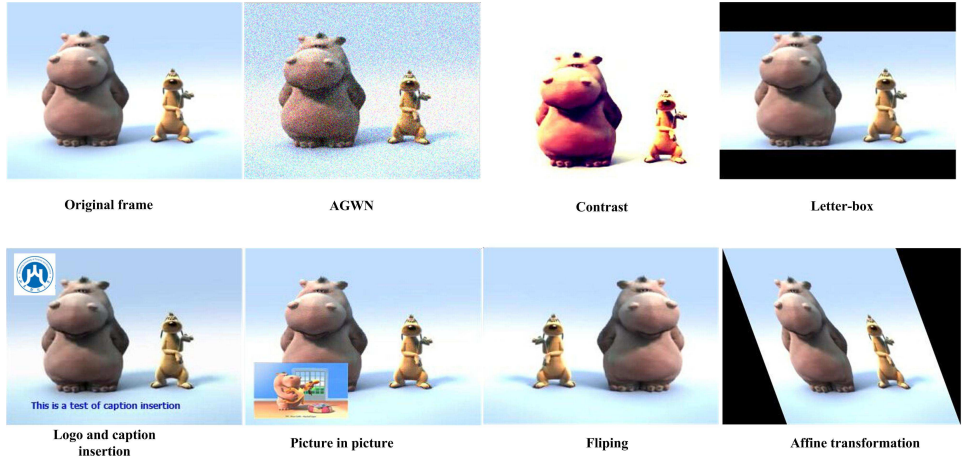}
\caption{The modifications on the original frame}
\label{corticalarchitecturefig}
\end{figure*}

\begin{table*}[htbp]
\centering \caption{Description of video attacks}
\footnotesize
\begin{tabular}{cc}
\hline\hline
Attacks  & Parameter Setting \\
 \hline
Rotation & 5 deg counterclockwise \\
AGWN & $\sigma_N$=110 \\
Blurring & motion blur with 10 pixels \\
Contrast enhancement & 1\% of data is saturated at low and high intensities \\
Letter-box & 10\% of pixels are replaced by black box at the top and bottom of frame\\
caption insertion & insert a line of text at the bottom of frame\\
Picture in picture & insert a different picture with size 100 x 100 in frame\\
Frame cropping & about 25\% cropping\\
Frame re-sampling & about 5\% frames changing\\
Affine transformation &transformation matrix [1 0 0; 0.5 1 0; 0 0 1]\\
\hline
\end{tabular}
\end{table*}

\section{Experimental Results and Analysis}
\subsection{Experimental Setting}
In the experiment, as mentioned, we take the global feature (normalized 64-bin histograms), local feature (SURF points) and temporal feature (the differences
of 64-bin histograms between adjacent frames) as the inputs of the scheme. Extensive experiments were conducted to evaluate the performance of the proposed
scheme. The original videos were downloaded from the CC\_WEB\_VIDEO Dataset (vireo.cs.cityu.edu.hk/webvideo) and OV Dataset (www.open-video.org), and then we apply different modifications on each video. Thus, almost 20,000 videos are used in
 our database for the experiments. A total of 11 single-mode modifications/attacks were applied on the test videos: (1) rotation; (2) Additive Gaussian White Noise (AGWN); (3) blurring;
 (4) contrast enhancement; (5) letter-box; (6) logo/caption insertion; (7) cropping; (8) flipping; (9) picture in picture; (10) affine transformation;
  and (11) re-sampling. Some of the modifications are shown in Fig. 6. The characterization or parameters of various modifications are provided in Table 1.
  Moreover, to evaluate the performance of the proposed scheme under combined modifications, we applied four types of combinations of more than one modification
  according to the ones defined by TRECVID [31], such as: (1) decrease in quality 1 (contrast+change of gamma+AGWN); (2) decrease in quality 2 (contrast+change of gamma+AGWN+blur+frame dropping);
  (3) post-production 1 (crop+flip+insertion of patterns); and (4)post-production 2 (crop+flip+insertion of patterns+picture in picture+shift).

We compared the proposed scheme with the LRTA method [23, 24]. Simultaneously, comparisons were also made to two classical algorithms, labeled the CGO [11], and the 3-D DCT [6] methods.

LRTA [23, 24]:  This method proposed a tensor-based hashing method (also called LRTA), and used spatial pixel values as feature, which is a single feature type
along the time axis, to generate the tensor. This method is not good at resisting local video editing, such as caption/logo insertion, picture in picture,
letter-box and shifting, which are common manipulations in UGC sites. They implement the tensor decomposition by CP model which has certain limitations.

CGO [11]: This method first divides a video frame into blocks and then combines direction information from pixels within the block into a single average or centroid statistic named CGO. Finally, the centroid statistics from each video frame are concatenated to form the final video fingerprint.

3-D DCT [6]: This method takes the video as a 3D sequence and uses DCT to generate a fingerprint vector.
\subsection{Performance Evaluation}
To evaluate the performance of the proposed scheme, quantitative and statistical evaluations are shown in this section, where F-score and the receiver operating characteristic (ROC) curve are used, respectively.
\begin{figure*}[htb!]
\begin{center}
  \subfigure[] {\scalebox{0.3}{\includegraphics[width=\textwidth,height=4.2in]{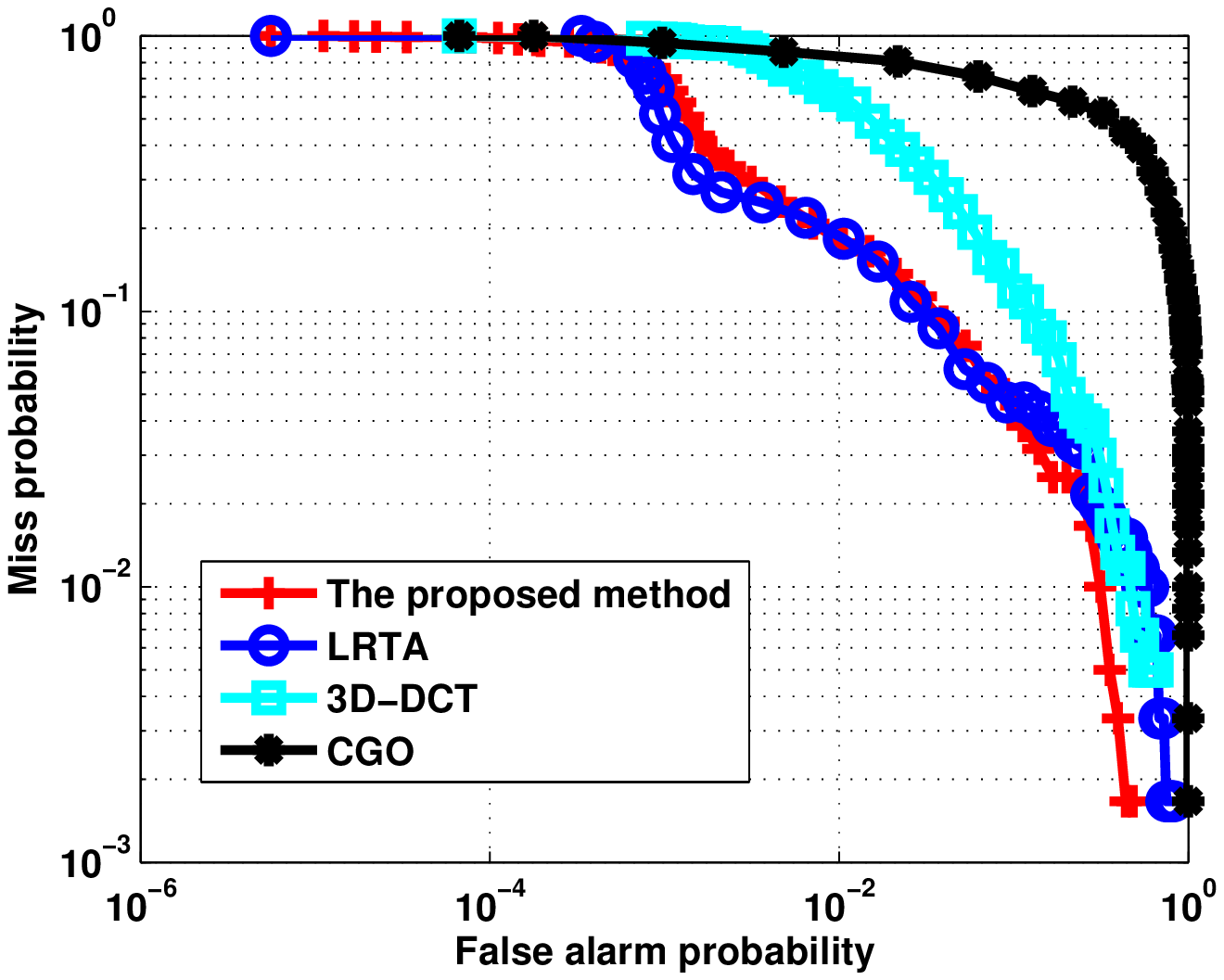}}}
  \subfigure[] {\scalebox{0.3}{\includegraphics[width=\textwidth,height=4.2in]{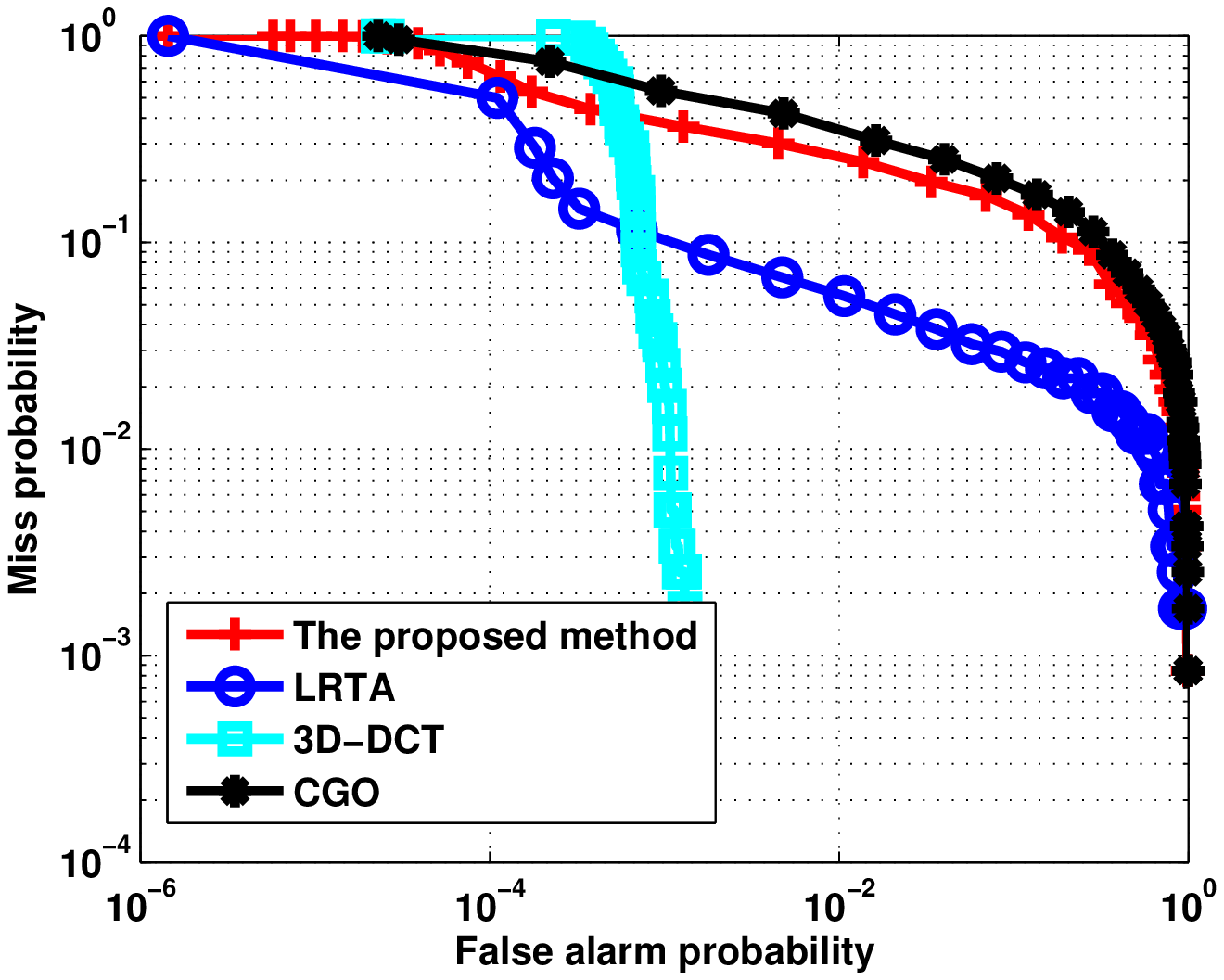}}}
  \subfigure[] {\scalebox{0.3}{\includegraphics[width=\textwidth,height=4.2in]{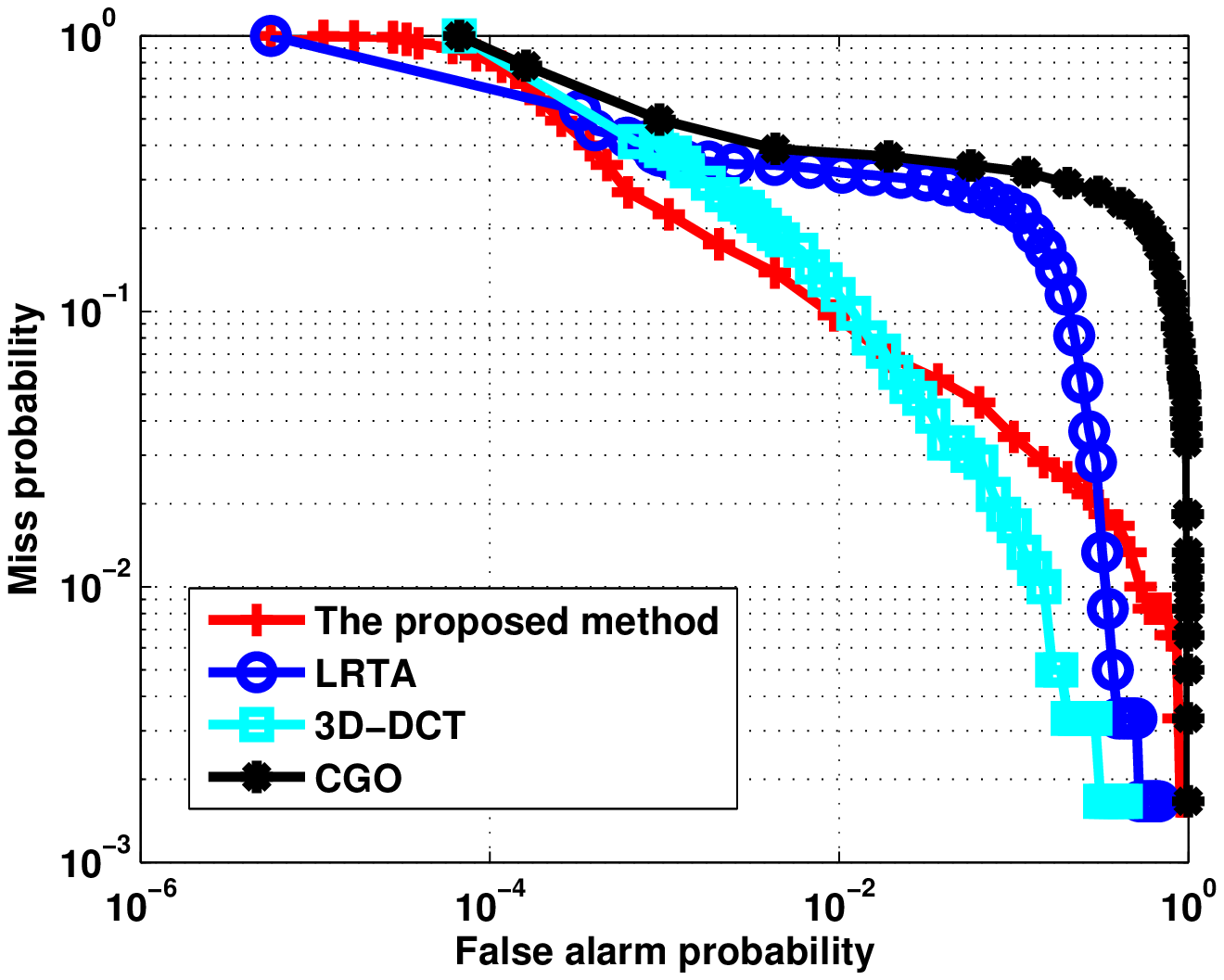}}}
  \subfigure[] {\scalebox{0.3}{\includegraphics[width=\textwidth,height=4.2in]{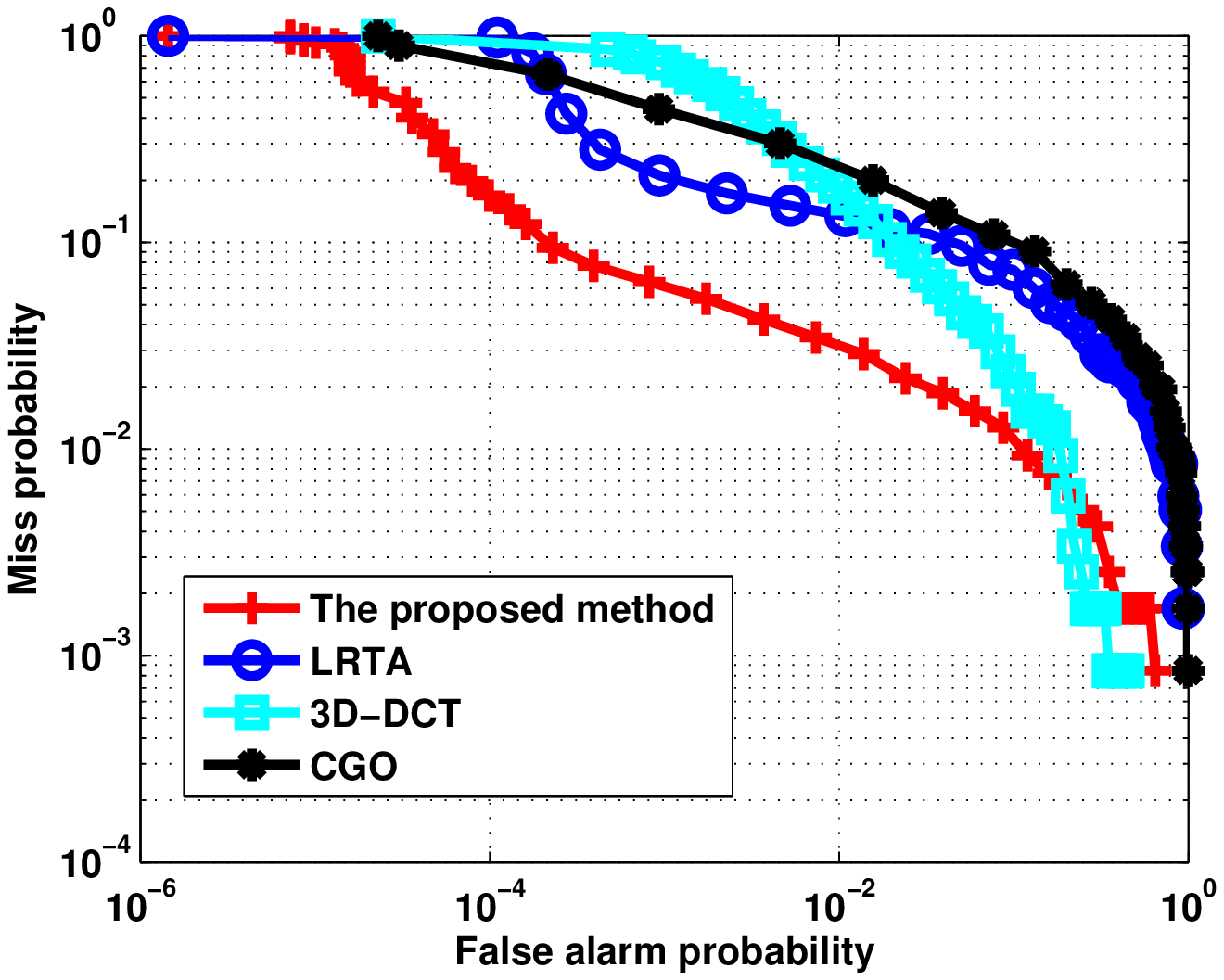}}}
  \subfigure[] {\scalebox{0.3}{\includegraphics[width=\textwidth,height=4.2in]{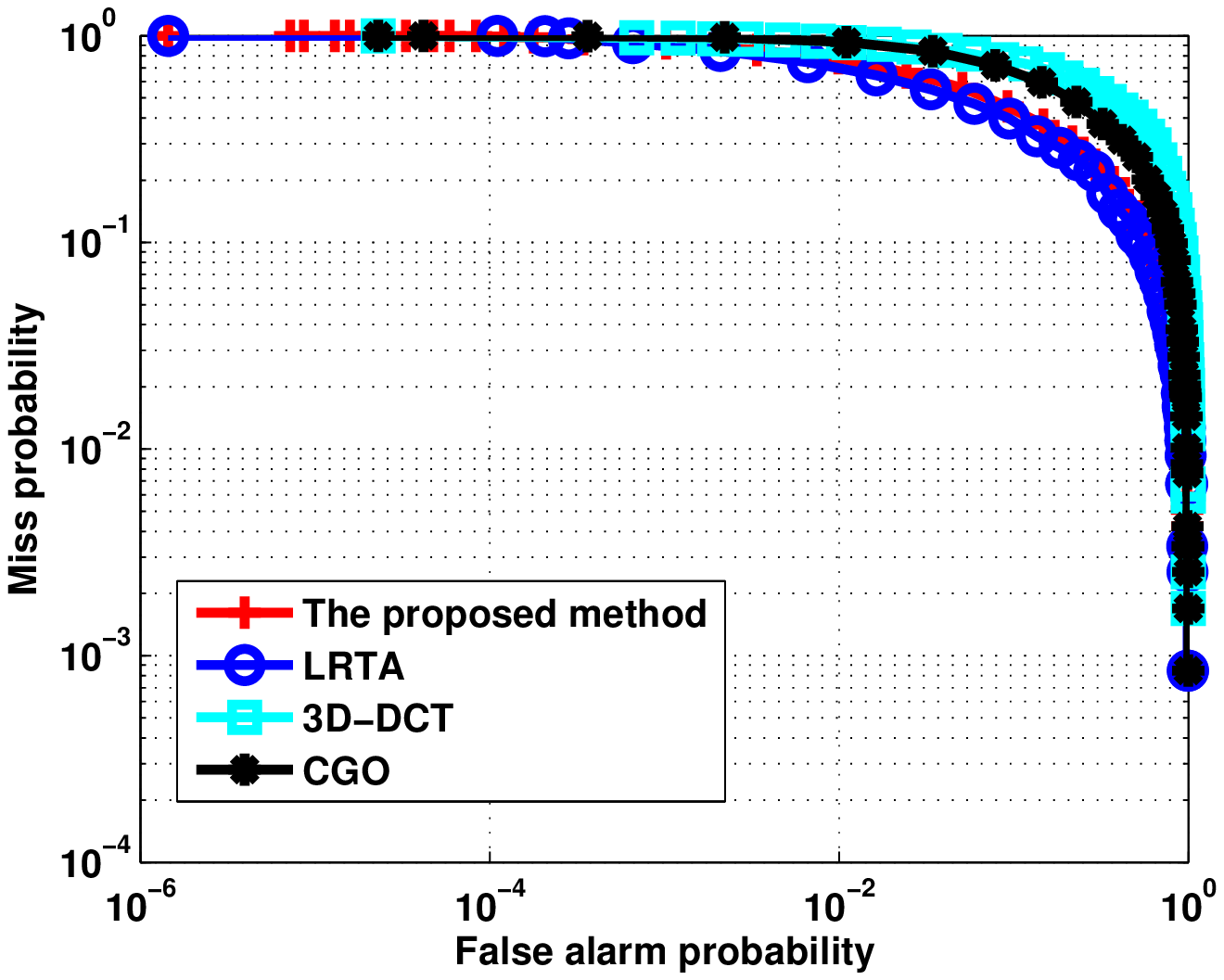}}}
  \subfigure[] {\scalebox{0.3}{\includegraphics[width=\textwidth,height=4.2in]{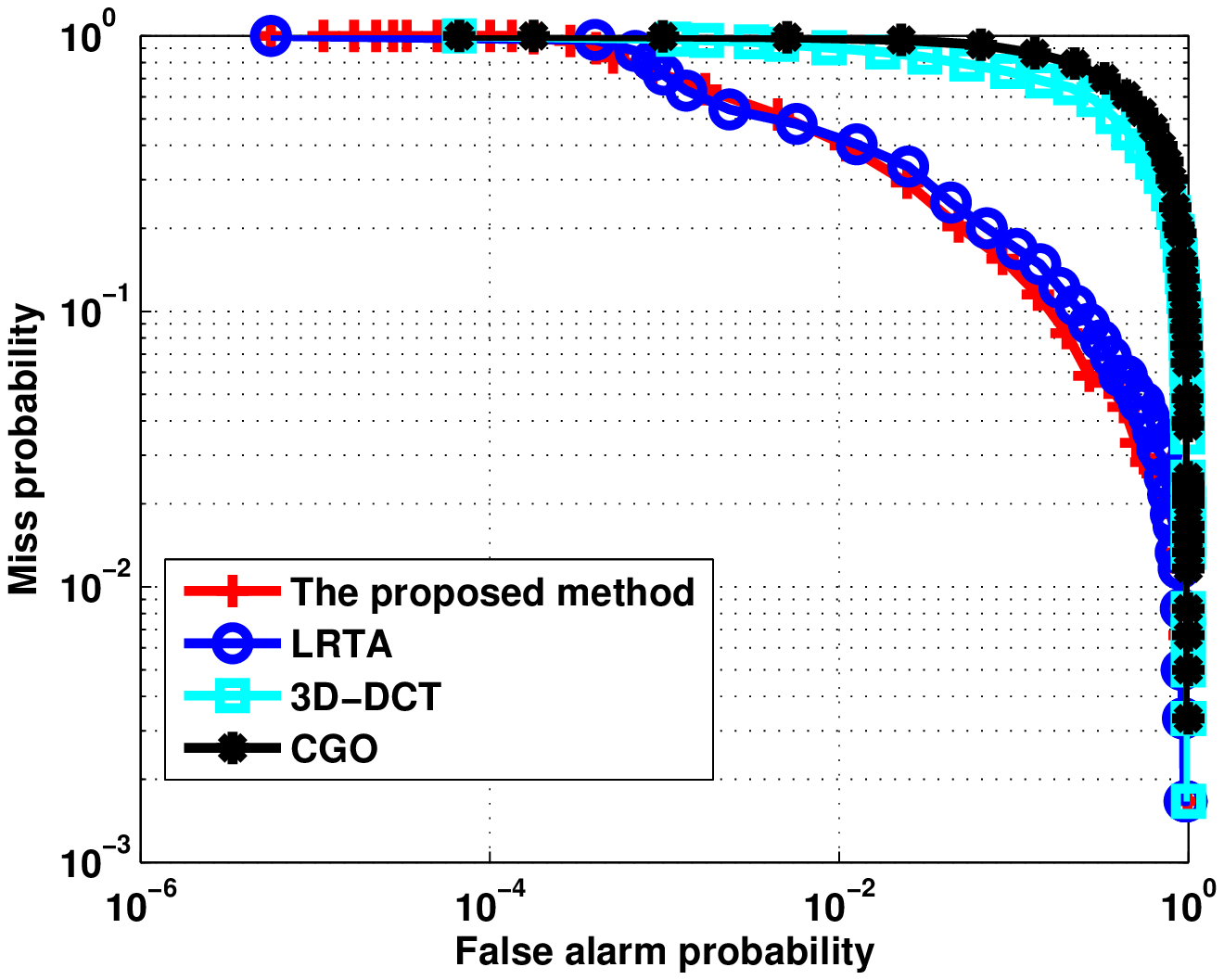}}}
  \subfigure[] {\scalebox{0.3}{\includegraphics[width=\textwidth,height=4.2in]{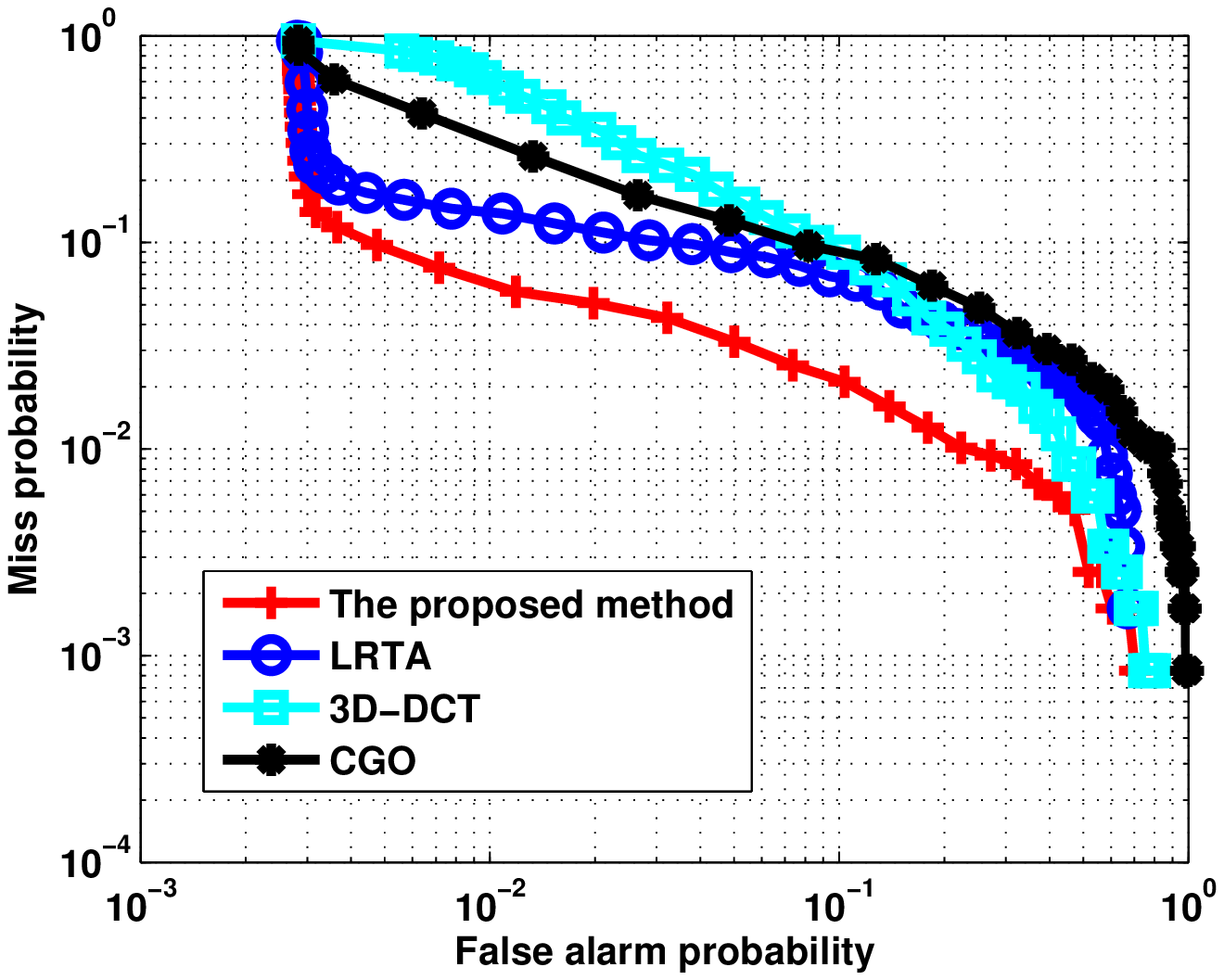}}}
  \subfigure[] {\scalebox{0.3}{\includegraphics[width=\textwidth,height=4.2in]{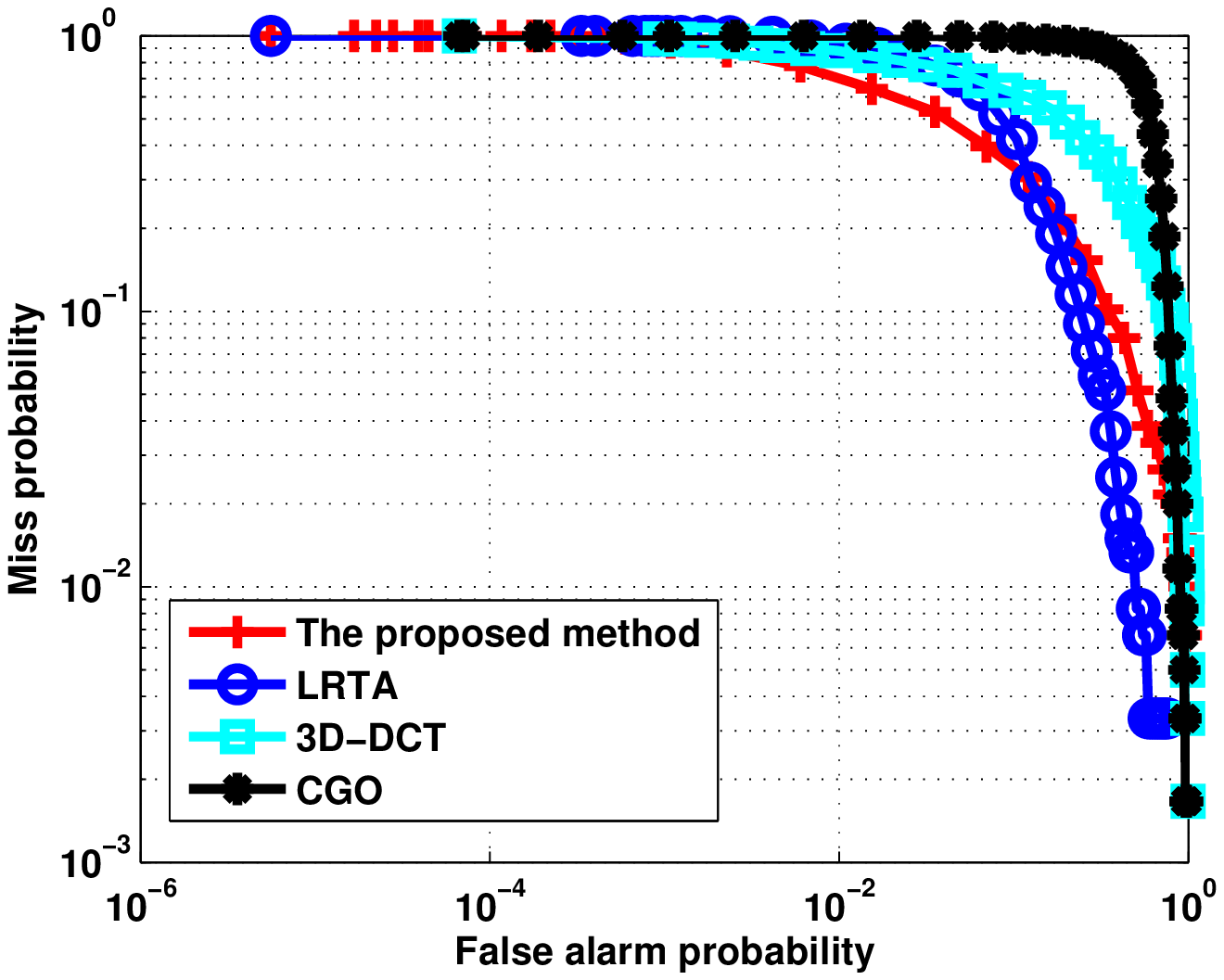}}}
  \subfigure[] {\scalebox{0.3}{\includegraphics[width=\textwidth,height=4.2in]{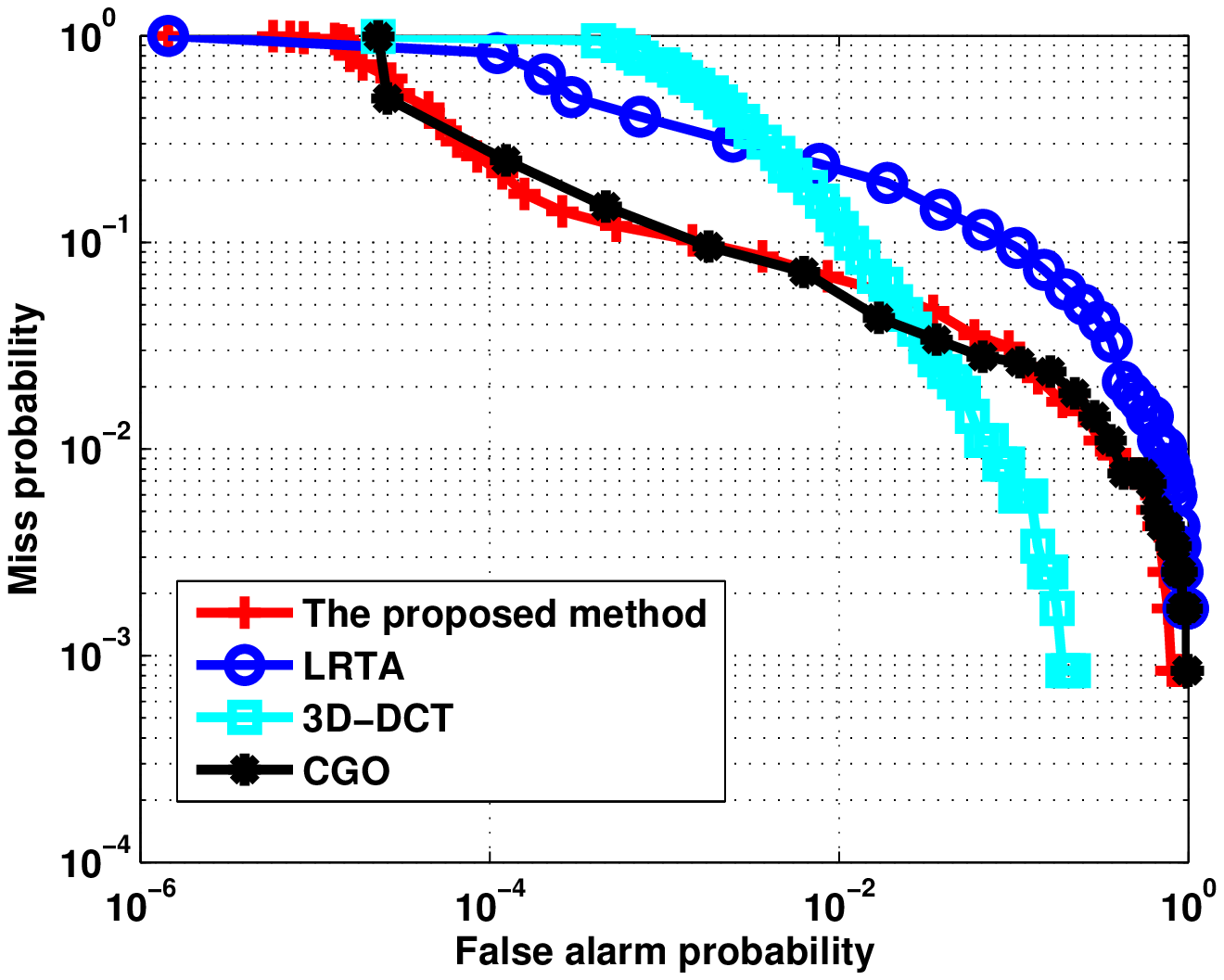}}}
  \subfigure[] {\scalebox{0.3}{\includegraphics[width=\textwidth,height=4.2in]{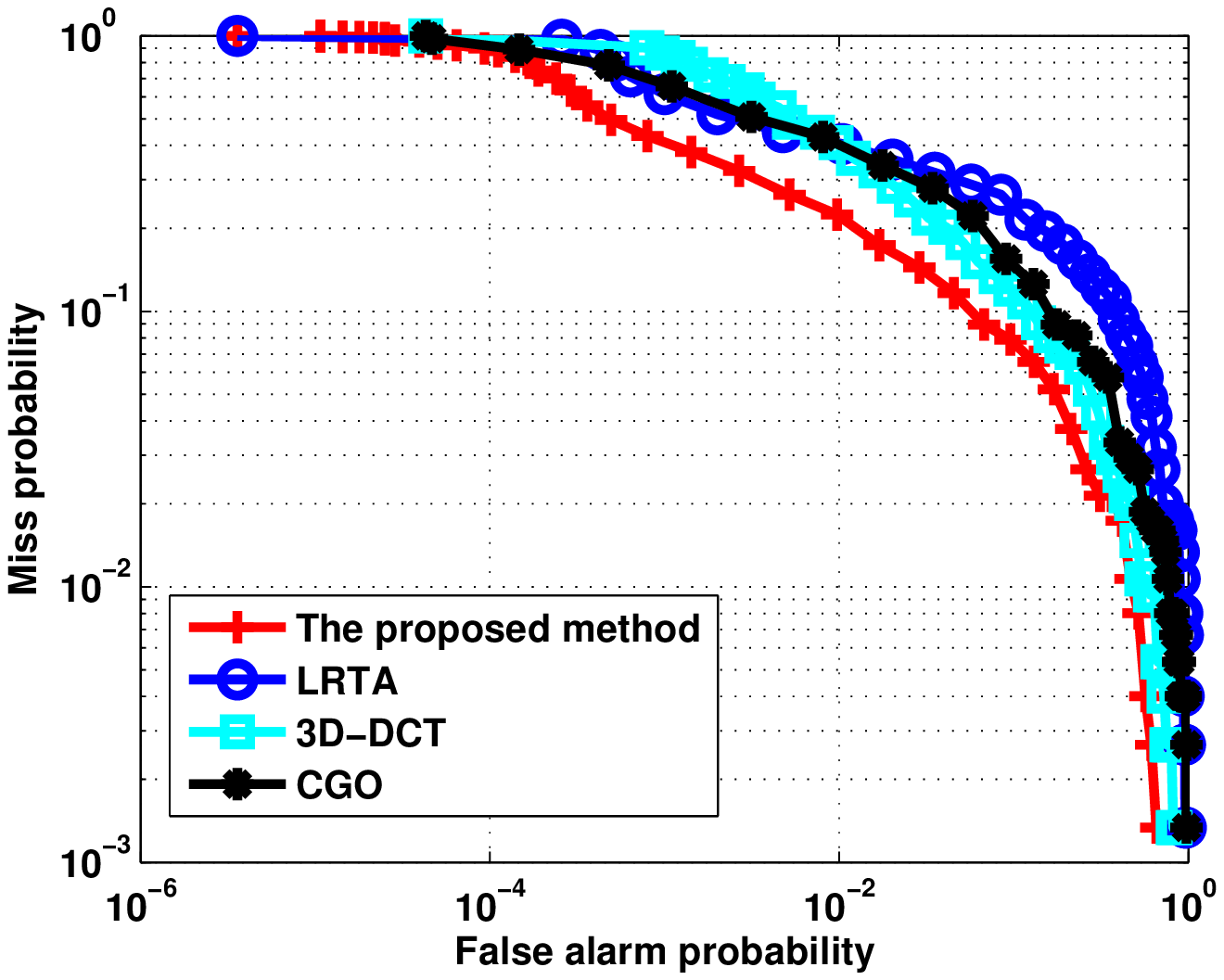}}}
   \subfigure[] {\scalebox{0.3}{\includegraphics[width=\textwidth,height=4.2in]{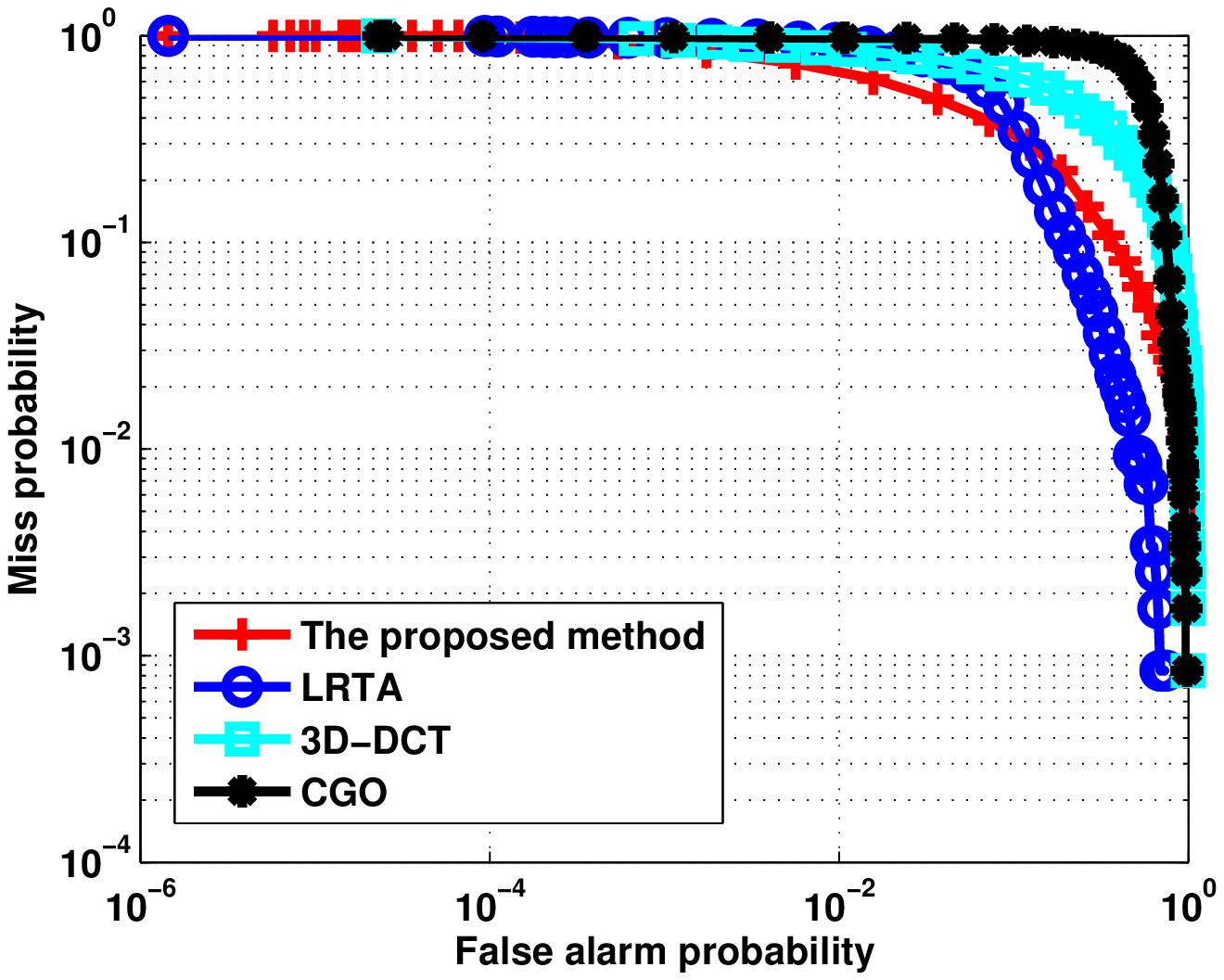}}}
   \subfigure[] {\scalebox{0.3}{\includegraphics[width=\textwidth,height=4.2in]{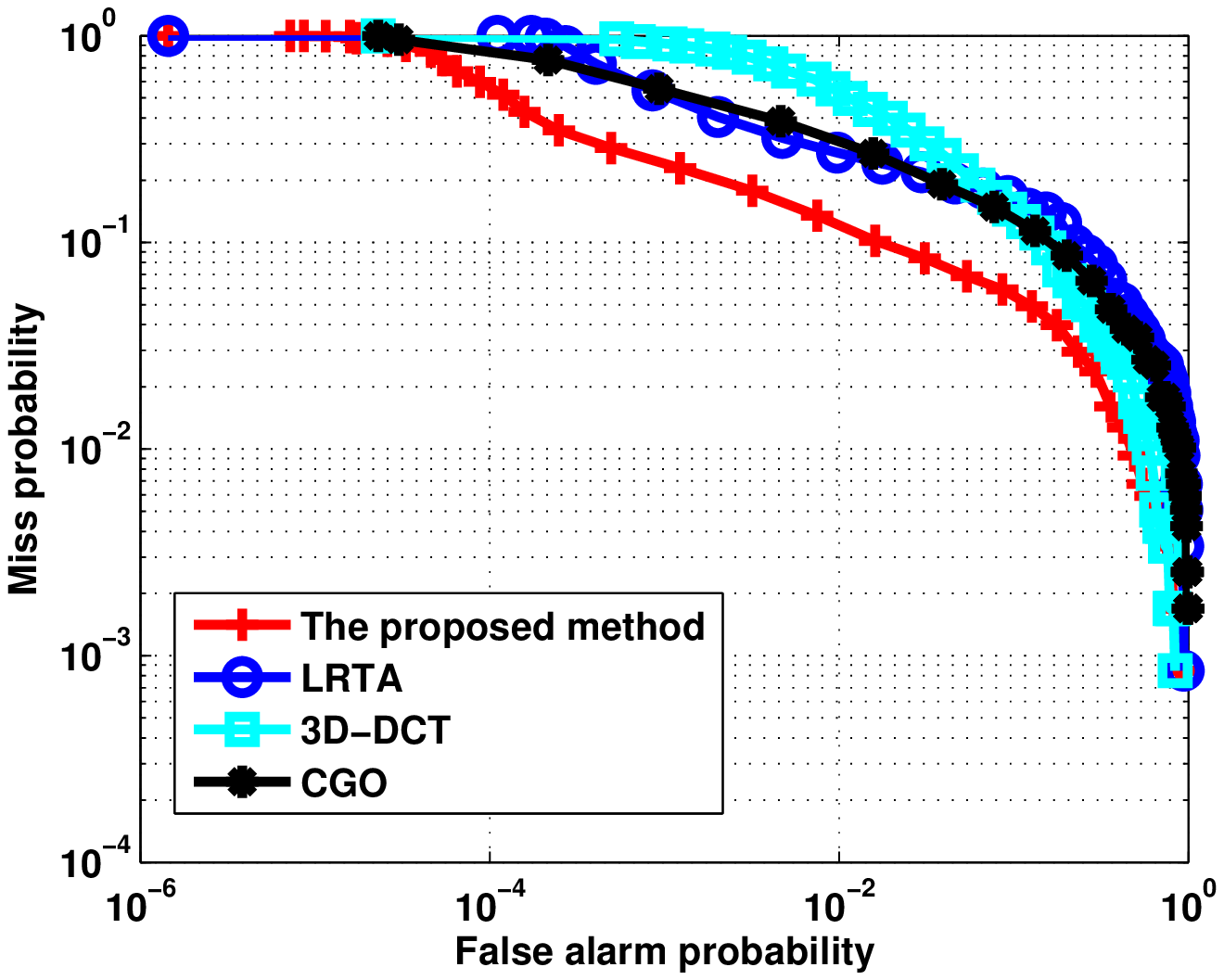}}}
   \subfigure[] {\scalebox{0.3}{\includegraphics[width=\textwidth,height=4.2in]{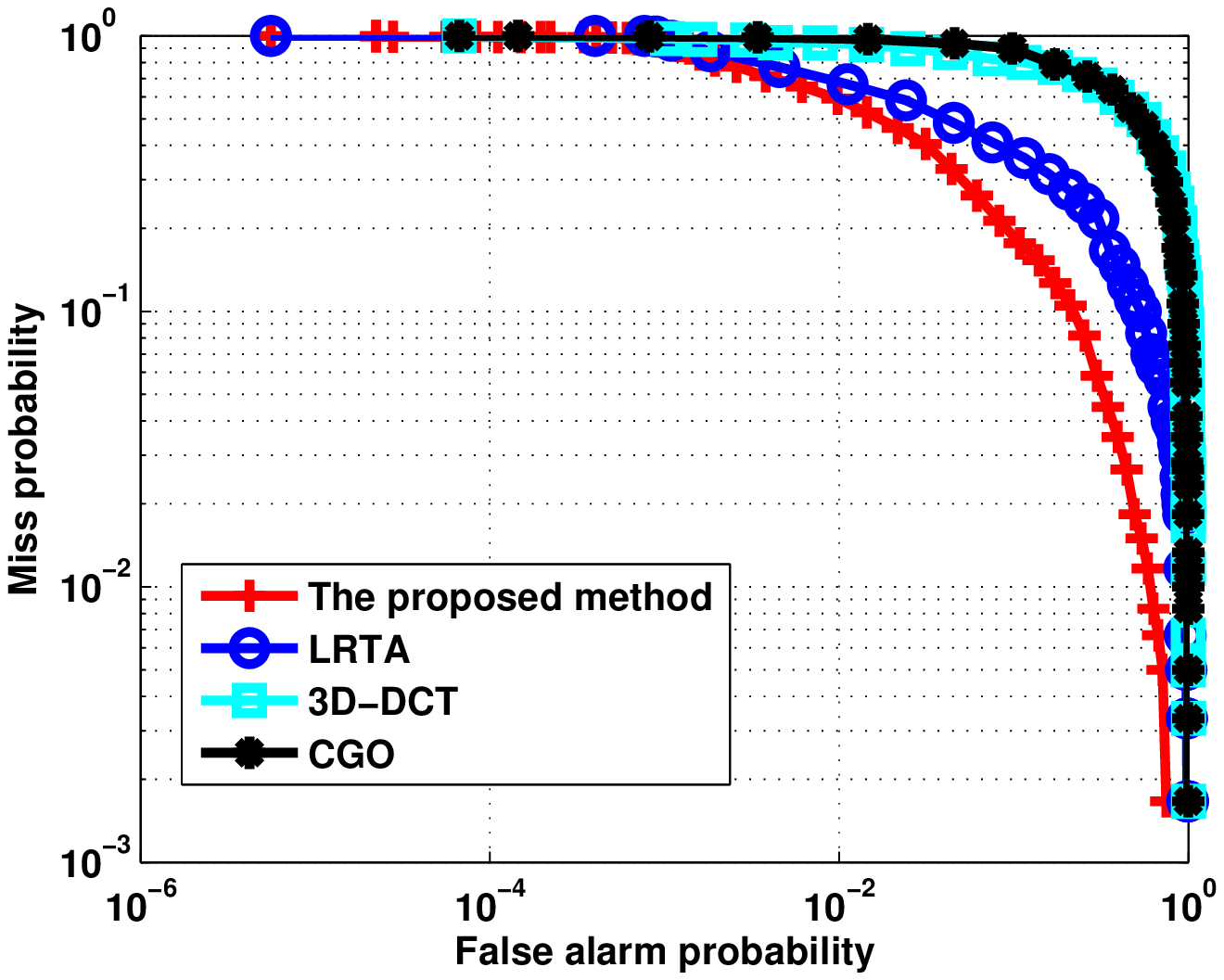}}}
   \subfigure[] {\scalebox{0.3}{\includegraphics[width=\textwidth,height=4.2in]{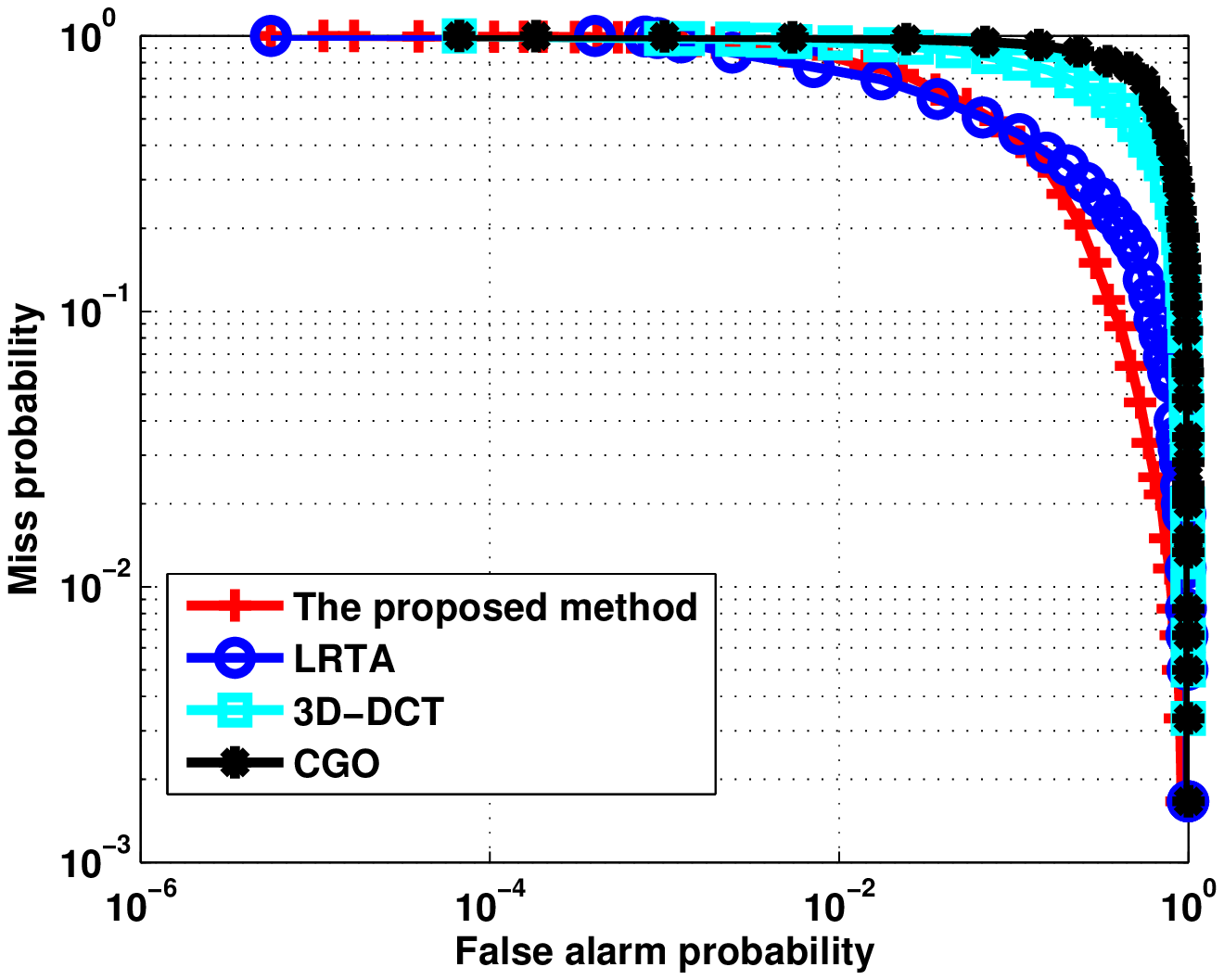}}}
   \subfigure[] {\scalebox{0.3}{\includegraphics[width=\textwidth,height=4.2in]{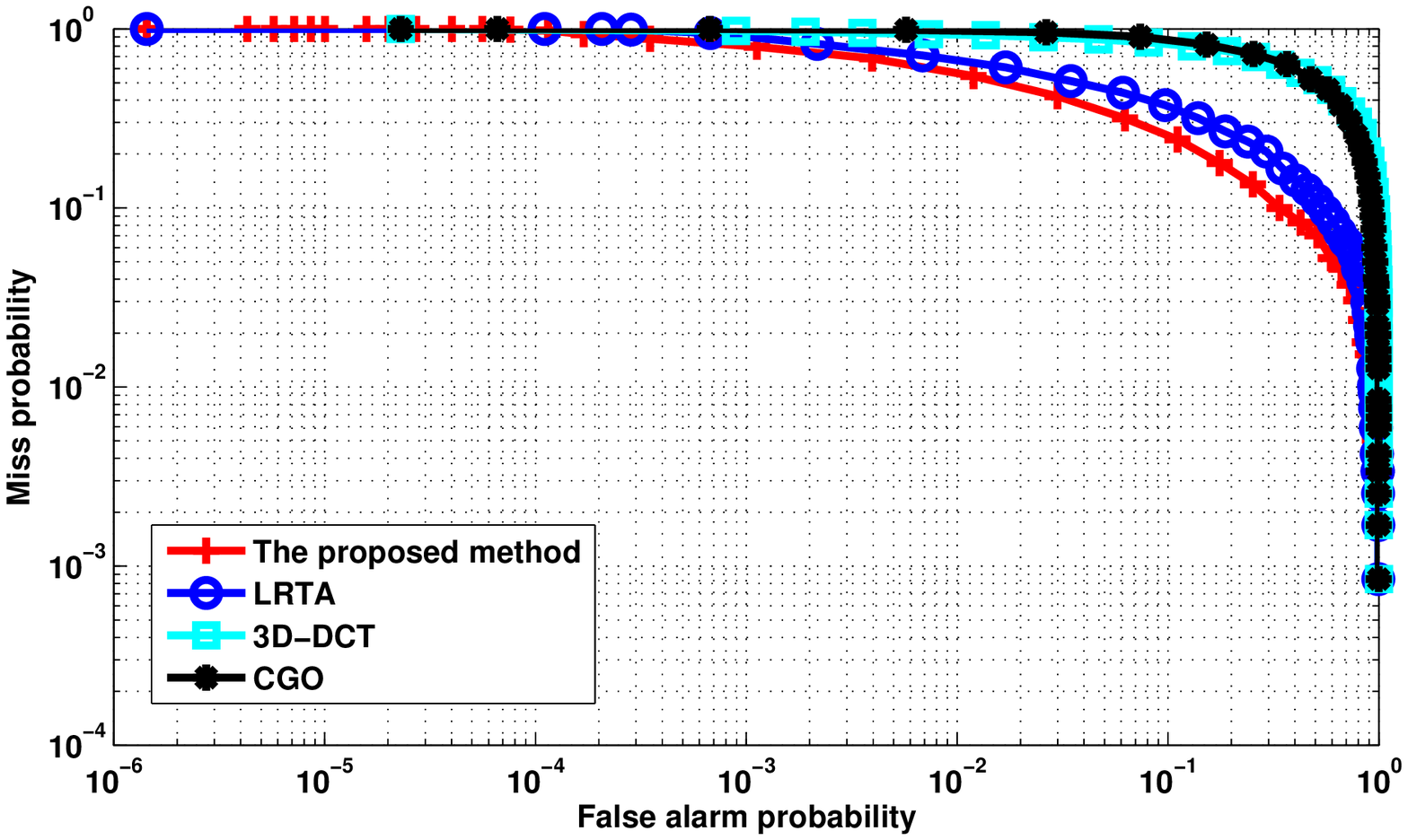}}}
  \caption{ The performance under different attacks: (a) Rotation; (b) AWGN; (c) Letter-box; (d) Caption insertion; (e) Cropping; (f) Flipping; (g) Picture in picture; (h) Affine transformation; (i) Contrast enhancement; (j) Logo substitution; (k)temporal resampling; (l) contrast+ change gamma+AGWN; (m) contrast+change of gamma+AGWN+blur+frame dropping; (n) crop+flip+insertion of patterns; (o) crop+flip+insertion of patterns+picture in picture+
shift}\label{inter_compare}
\end{center}
\end{figure*}

\begin{figure*}[htb!]
\begin{center}
  \subfigure[] {\scalebox{0.44}{\includegraphics[width=\textwidth,height=4.6in]{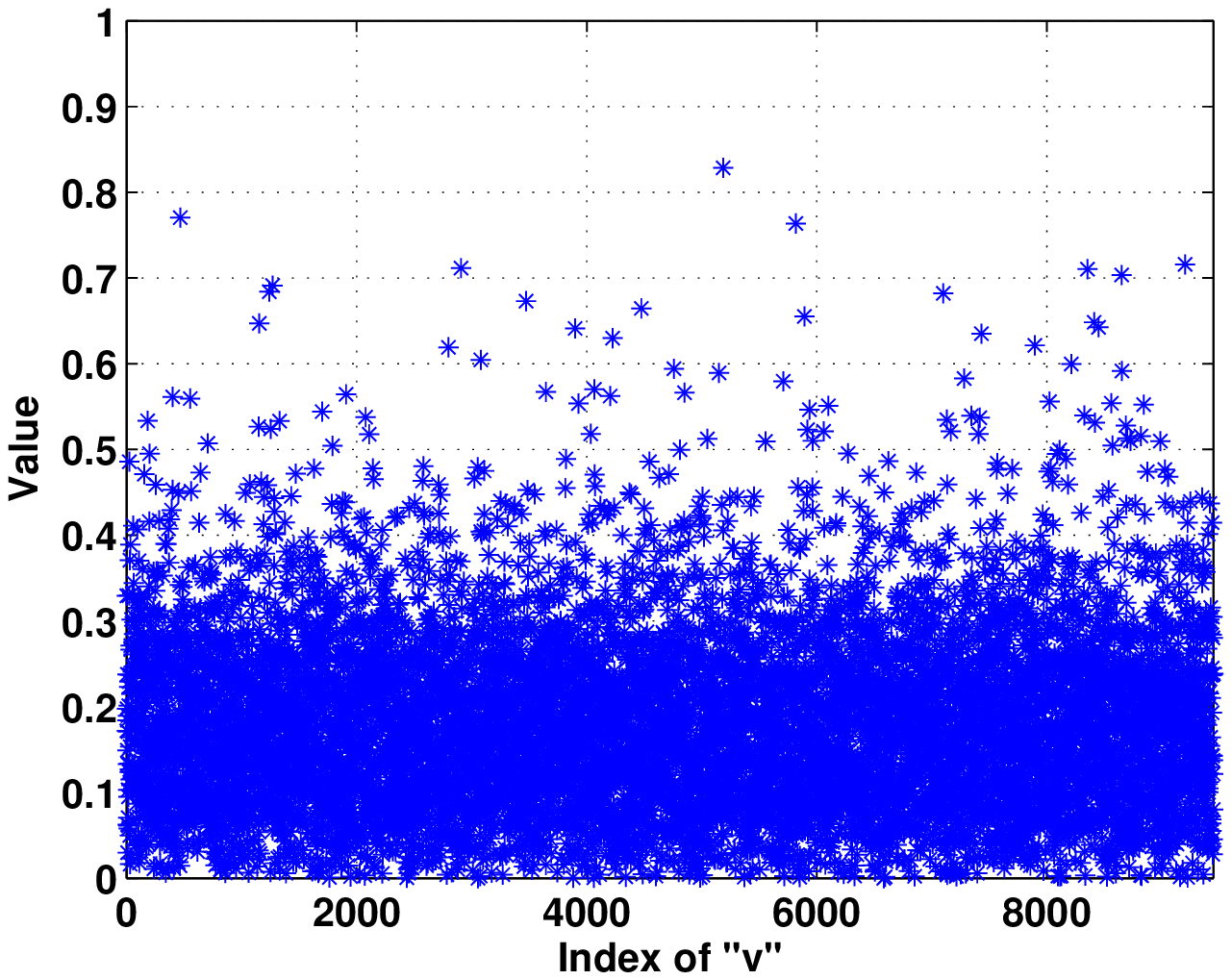}}}
  \subfigure[] {\scalebox{0.44}{\includegraphics[width=\textwidth,height=4.6in]{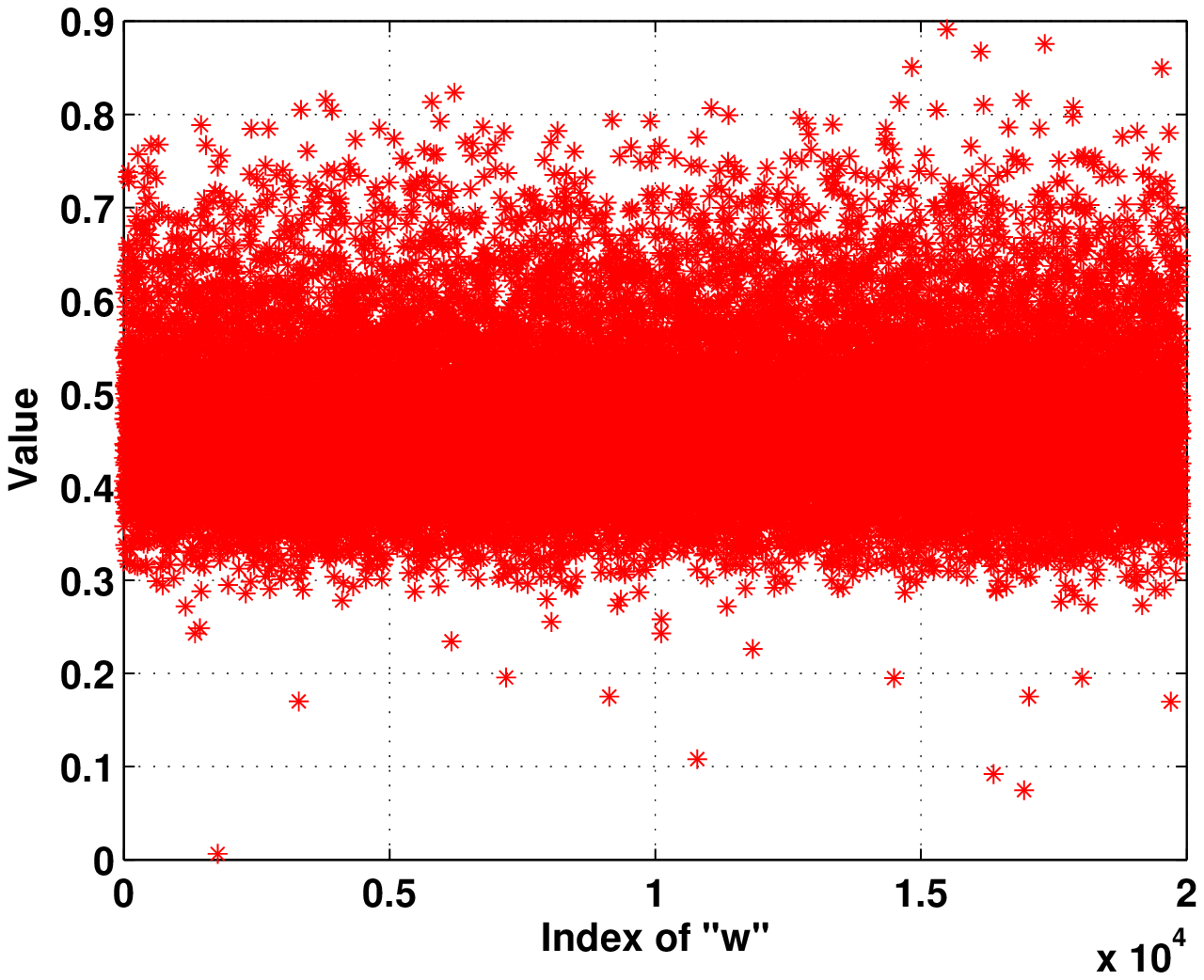}}}
  \subfigure[] {\scalebox{0.44}{\includegraphics[width=\textwidth,height=4.6in]{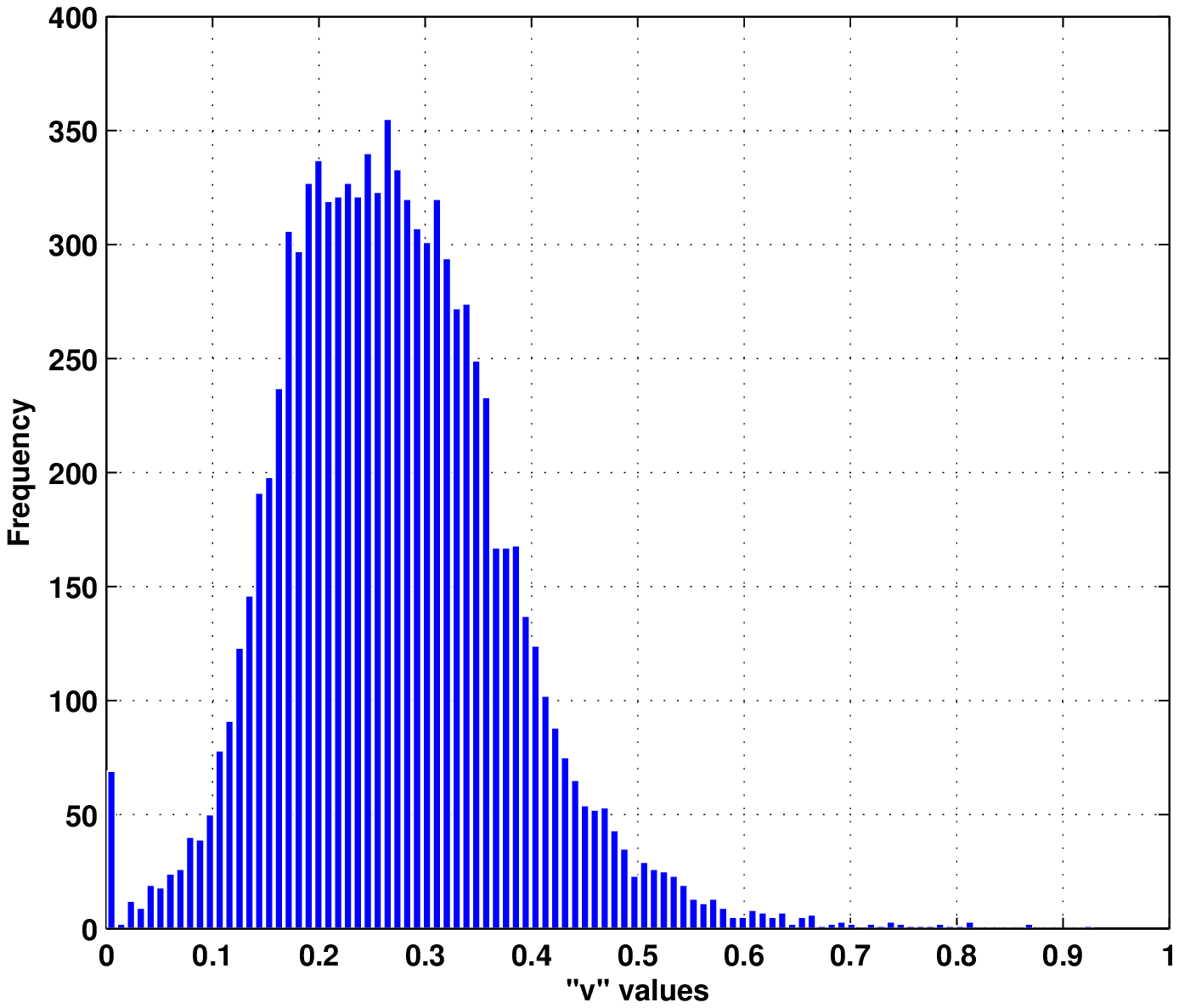}}}
  \subfigure[] {\scalebox{0.44}{\includegraphics[width=\textwidth,height=4.6in]{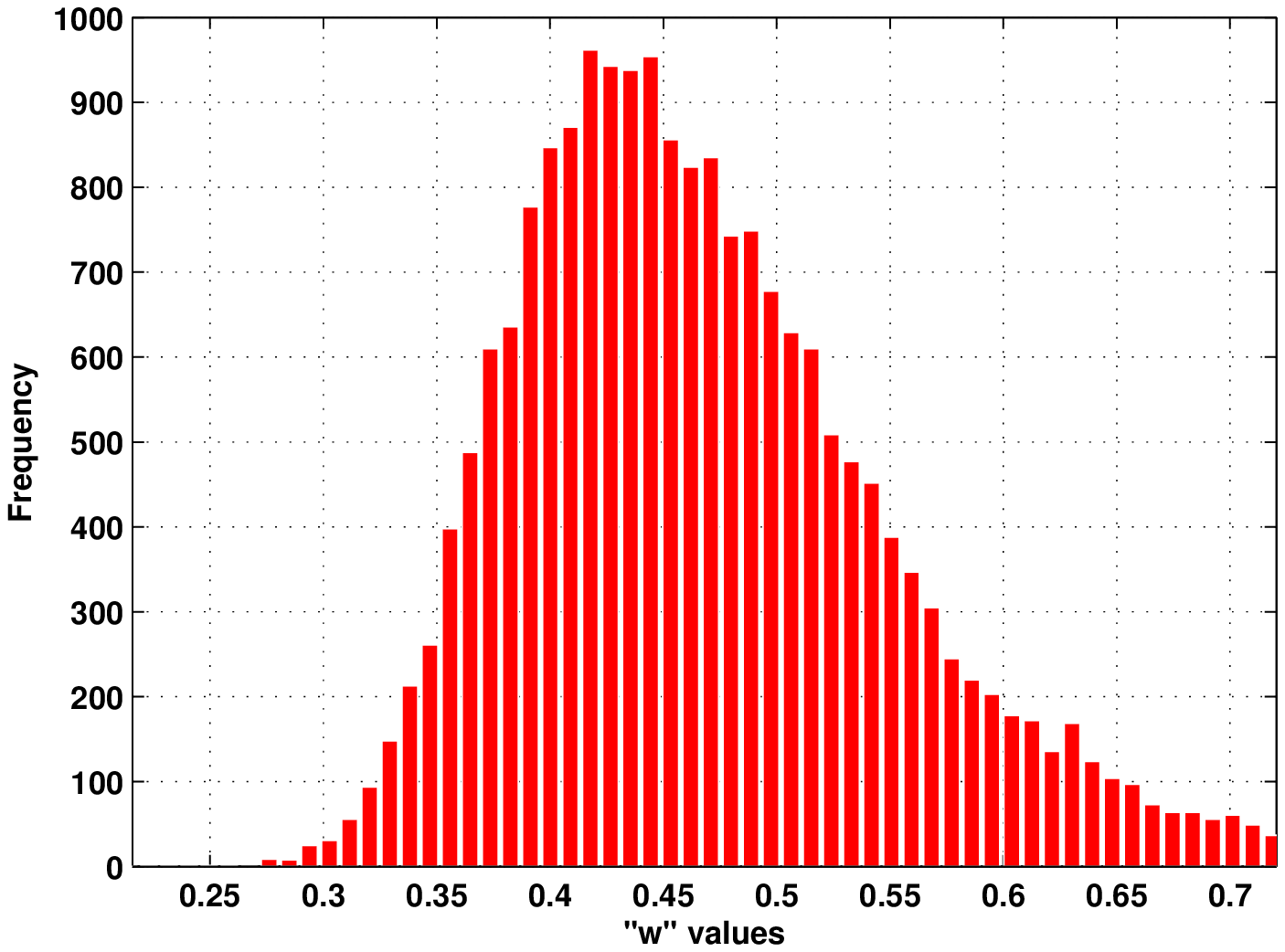}}}
  \caption{ The description of values $w$ and $v$: (a) Original value of $v$; (b)Original values of $w$; (c) bar of $v$; (d) bar of "w"}\label{inter_compare}
  \end{center}
\end{figure*}

\subsubsection{Quantitative Evaluation}
Robustness and discrimination are important for a video fingerprinting system. Generally, robustness and discrimination are expressed by recall and precision [32],
respectively. Recall is the same as the true positive rate (a measure of robustness of the system), whereas precision is a measure of discrimination and is
defined as the percentage of correct hits within all detected copies. Based on recall and precision, F-score which is a single combined metric is taken as a
quantitative measure in this study, and is defined as follows [32]:
\begin{equation}
{F_\beta } = (1 + {\beta ^2})\frac{{{P_p} * {P_r}}}{{{\beta ^2}{P_p} + {P_r}}}
\end{equation}
where ${P_p}$ and ${P_r}$ are precision and recall rate. $\beta$ is a parameter that defines how much weight should be given to recall versus precision. F-score
is between 0 and 1, with 1 representing a perfect performance that is completely robust and completely discriminant (100\% precision and 100\% recall).
A low F-score close to 0 represents a poor system in terms of both robustness and discrimination. We used $\beta=0.5$ in the experiments, which provides twice
as much importance to precision as to recall, because a video fingerprinting system should have high precision to minimize the amount of human interaction required [32].
A larger F-score means better performance of the algorithm. The F-score results under various modifications are shown in Table 2. In contrast to other methods,
the proposed method is better under most modifications. The improvement under combined modification is larger than the ones under single-mode modification.
\begin{table}[htbp]
\centering \caption{F-score results}
\footnotesize
\begin{tabular}{ccccc}
\hline\hline
Modifications  & Proposed  & LRTA & 3D-DCT & CGO \\
 \hline
Letter-box & \textbf{0.9715} &  0.9253  & 0.9708    & 0.9112 \\
Logo insertion & \textbf{0.9476}    &0.8908 &0.9234 &0.9046 \\
Noise   &0.9294 &0.9820 &\textbf{0.9989}    &0.9083\\
caption insertion   &\textbf{0.9883}    &0.9618 &0.9627 &0.9411\\
Contrast    &\textbf{0.9788}    &0.9403 &0.9731 &0.9784\\
Picture in picture  &\textbf{0.9786}    &0.9612 &0.9251 &0.9411\\
Cropping    &0.8015 &\textbf{0.8226 }&0.6527    &0.7036\\
Flipping    &\textbf{0.9063}    &0.9051 &0.6403 &0.5464\\
Frame changing  &\textbf{0.8428}    &0.8423 &0.7148 &0.4680\\
Combined modification 1 &\textbf{0.9655}    &0.9279 &0.9097 &0.9263\\
Combined modification 2 &\textbf{0.8725}    &0.8292 &0.5745 &0.5556\\
\hline
\end{tabular}
\end{table}

\subsubsection{Statistical Evaluation}
The miss and false alarm probability are considered in the ROC curve. The miss probability is defined as the probability of true copies without detection,
whereas false alarm probability is defined as the probability of false positive copies that are actually negative cases. Fig. 7 shows the ROC (log) curves
under different attacks, and the performance of the proposed scheme is better than the other performance under almost all modifications, especially under some
popular manipulations in the user-generated videos and video post-production, such as caption insertion and picture-in-picture. In addition, the performance has
 significant improvement under the combined modifications, which can be observed in Fig. 7(l-o). However, the performance under AWGN is not as good as the other performances.
  The main reason is that some noisy points are wrongly considered as interested points in the SURF detector.

\subsection{Threshold Analysis}
In real application, the video fingerprinting system should decide whether the query video is a modified copy. The common method is to set a threshold $\tau $ in advance.
Based on the assumption that the original and query fingerprints are ${\bf{F}}(V)$ and ${\bf{F}}({V_A})$, respectively, a decision is made as follows.
\begin{equation}
\left\{ {\begin{array}{*{20}{c}}
{||{\bf{F}}(V) - {\bf{F}}({V_A})|{|_2} \le \tau }&{{V_A}{\kern 1pt} {\kern 1pt} is{\kern 1pt} {\kern 1pt} a{\kern 1pt} {\kern 1pt} copy{\kern 1pt} {\kern 1pt} of{\kern 1pt} {\kern 1pt} V}\\
{||{\bf{F}}(V) - {\bf{F}}({V_A})|{|_2} > \tau }&{{\kern 1pt} {\kern 1pt} {\kern 1pt} {V_A}{\kern 1pt} {\kern 1pt} and{\kern 1pt} {\kern 1pt} V{\kern 1pt} {\kern 1pt} are{\kern 1pt} {\kern 1pt} different}
\end{array}} \right.
\end{equation}

The threshold is important in a real video fingerprinting system. A smaller threshold can improve the true positive probability, but negatively affects the miss probability.
 By contrast, a larger threshold causes a lower miss probability, but the false alarm probability may be higher as a result. Therefore, the choice of threshold should be considered in a real system. To analyze the choice of this threshold, we first list some symbol notations in Table 3.

\begin{table}[htbp]
\newcommand{\tabincell}[2]{\begin{tabular}{@{}#1@{}}#2\end{tabular}}
\centering \caption{Notations}
\footnotesize
\begin{tabular}{cc}
\hline
\hline
Symbol &Defination\\
\hline
$w$   & \tabincell{c}{L2-norm of the fingerprint distance between two fingerprints\\of visually different video contents }  \\

$v$   & \tabincell{c}{L2-norm of the fingerprint distance between two fingerprints\\visually similar video content }  \\

$\tau$ &The threshold\\

$P( \bullet)$ & Distribution of $``\bullet"$\\
 \hline
\end{tabular}
\end{table}

\begin{figure}[htbp] \centering
\includegraphics [width=3in]{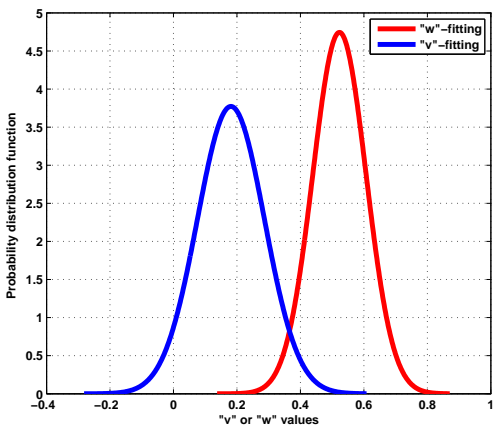}
\caption{The illustration of threshold selection}
\label{corticalarchitecturefig}
\end{figure}
\begin{figure*}[htb!]
\begin{center}
  \subfigure[] {\scalebox{0.22}{\includegraphics[width=\textwidth,height=4.6in]{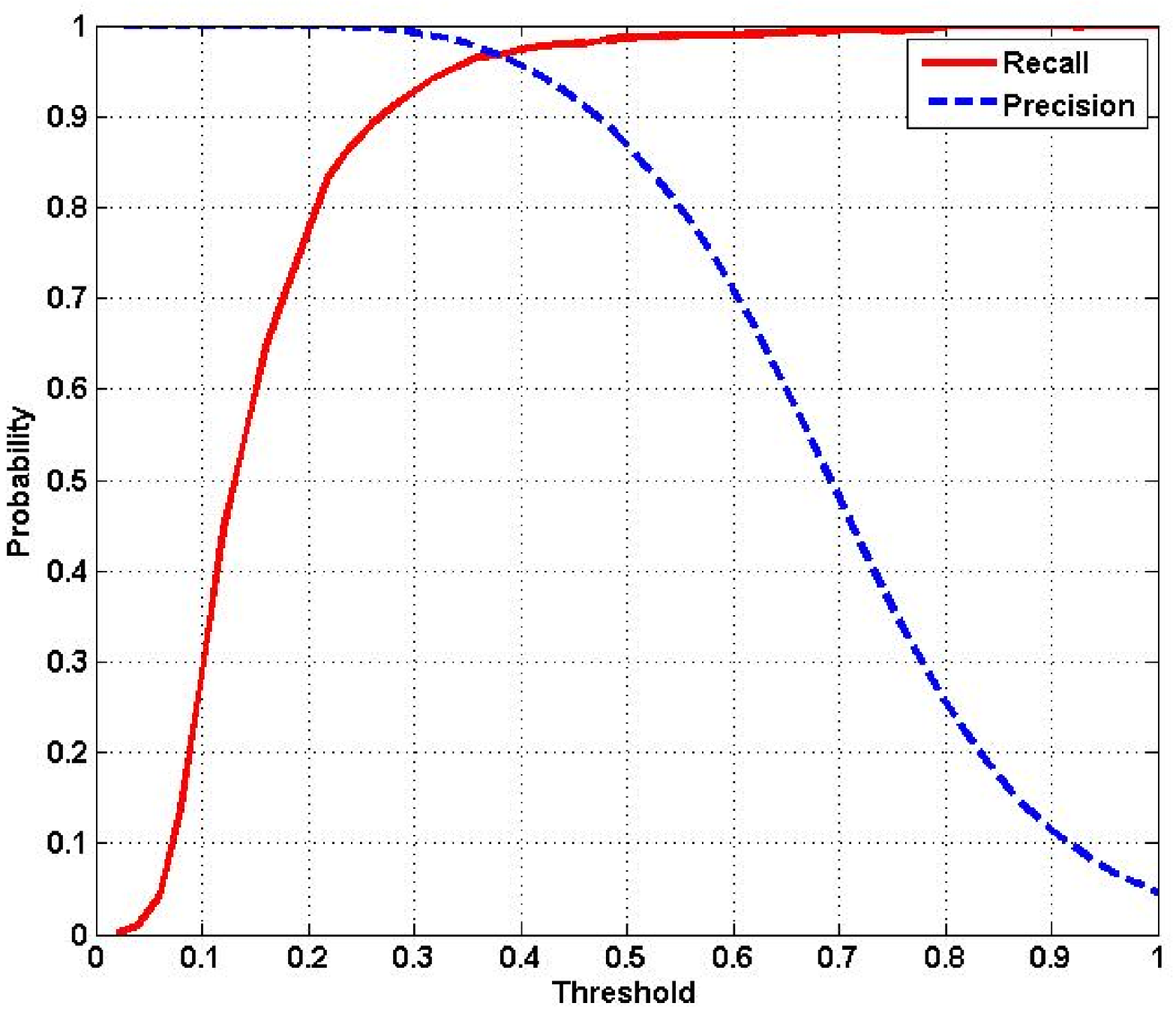}}}
  \subfigure[] {\scalebox{0.22}{\includegraphics[width=\textwidth,height=4.6in]{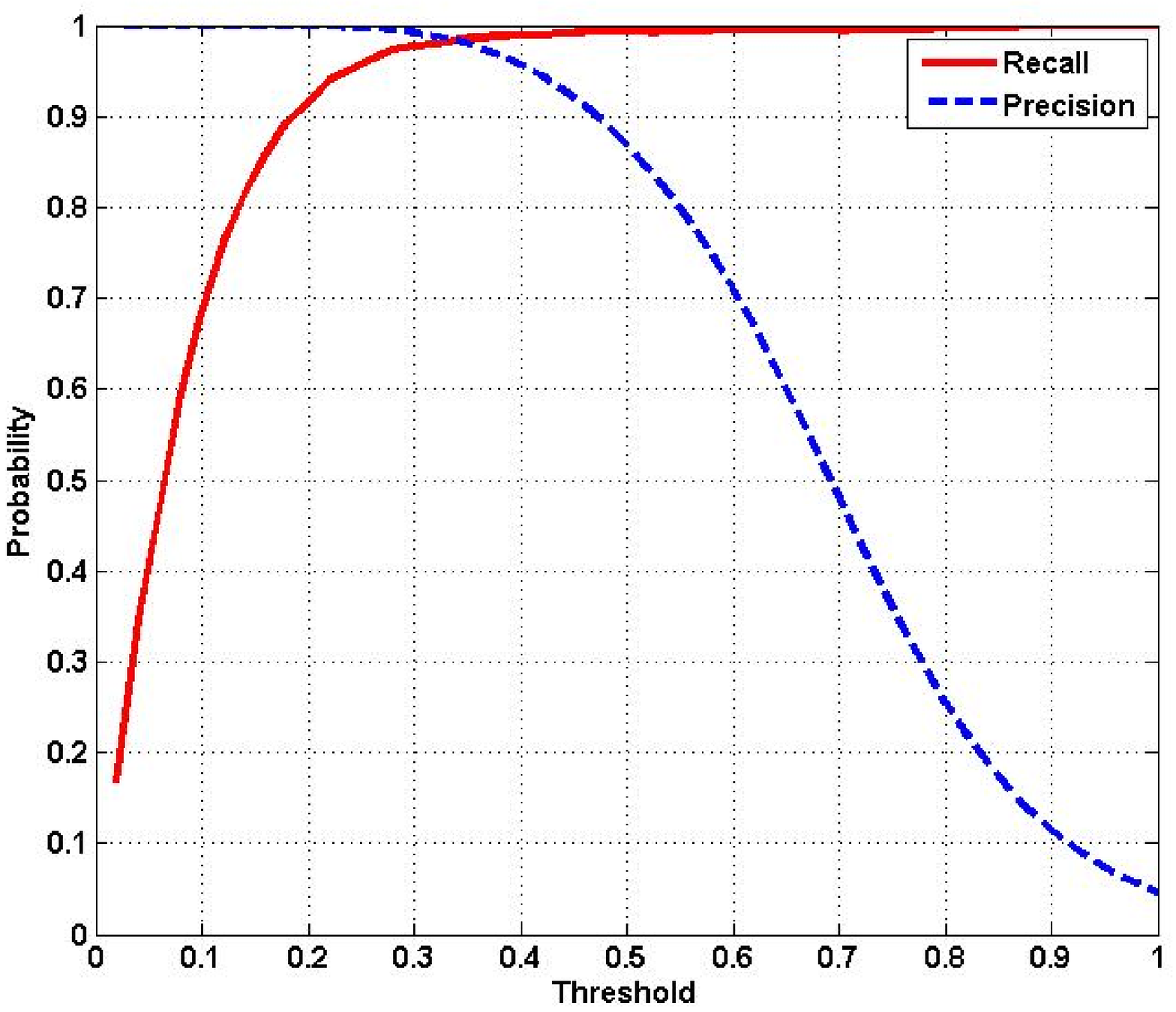}}}
  \subfigure[] {\scalebox{0.22}{\includegraphics[width=\textwidth,height=4.6in]{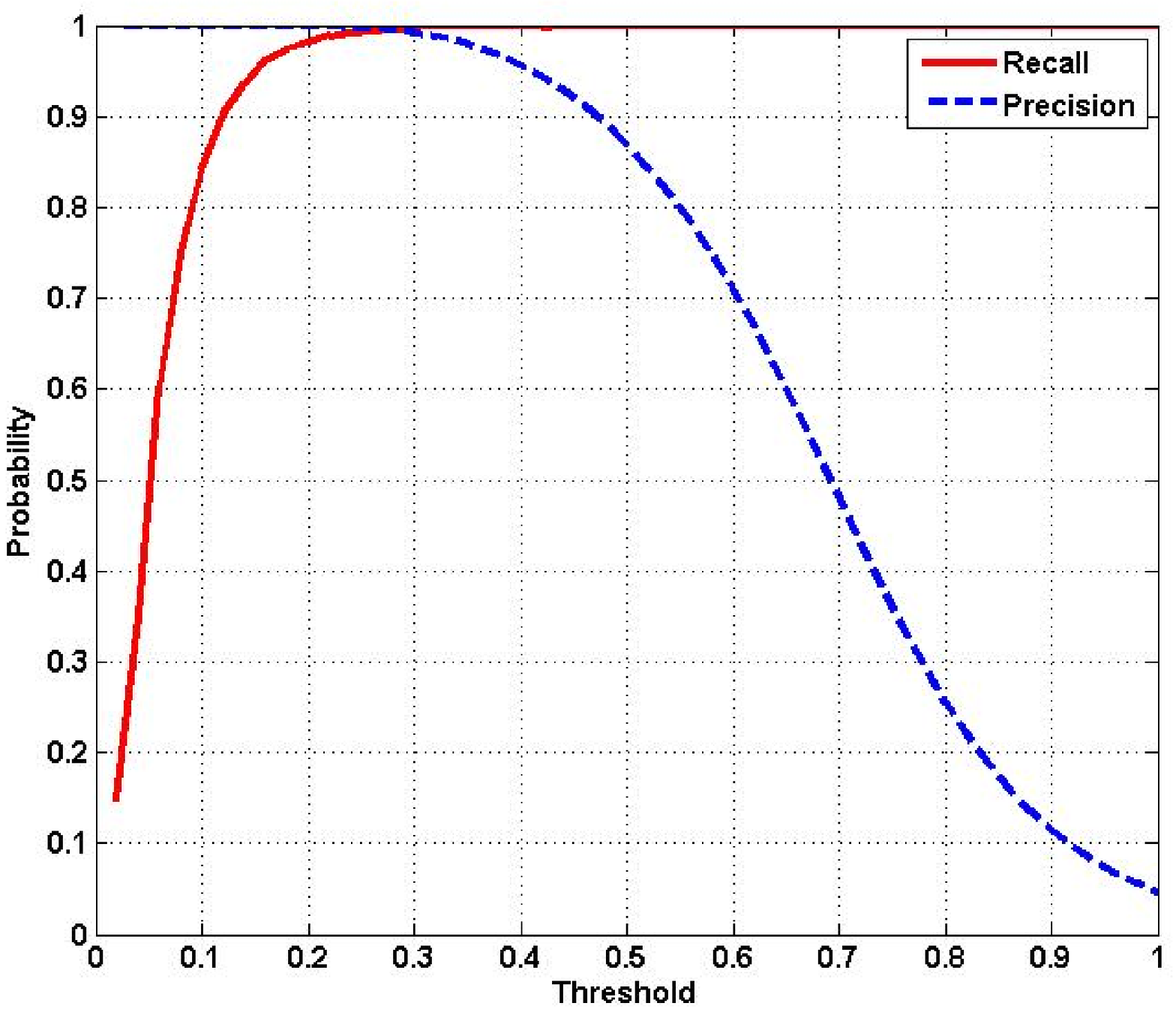}}}
  \subfigure[] {\scalebox{0.22}{\includegraphics[width=\textwidth,height=4.6in]{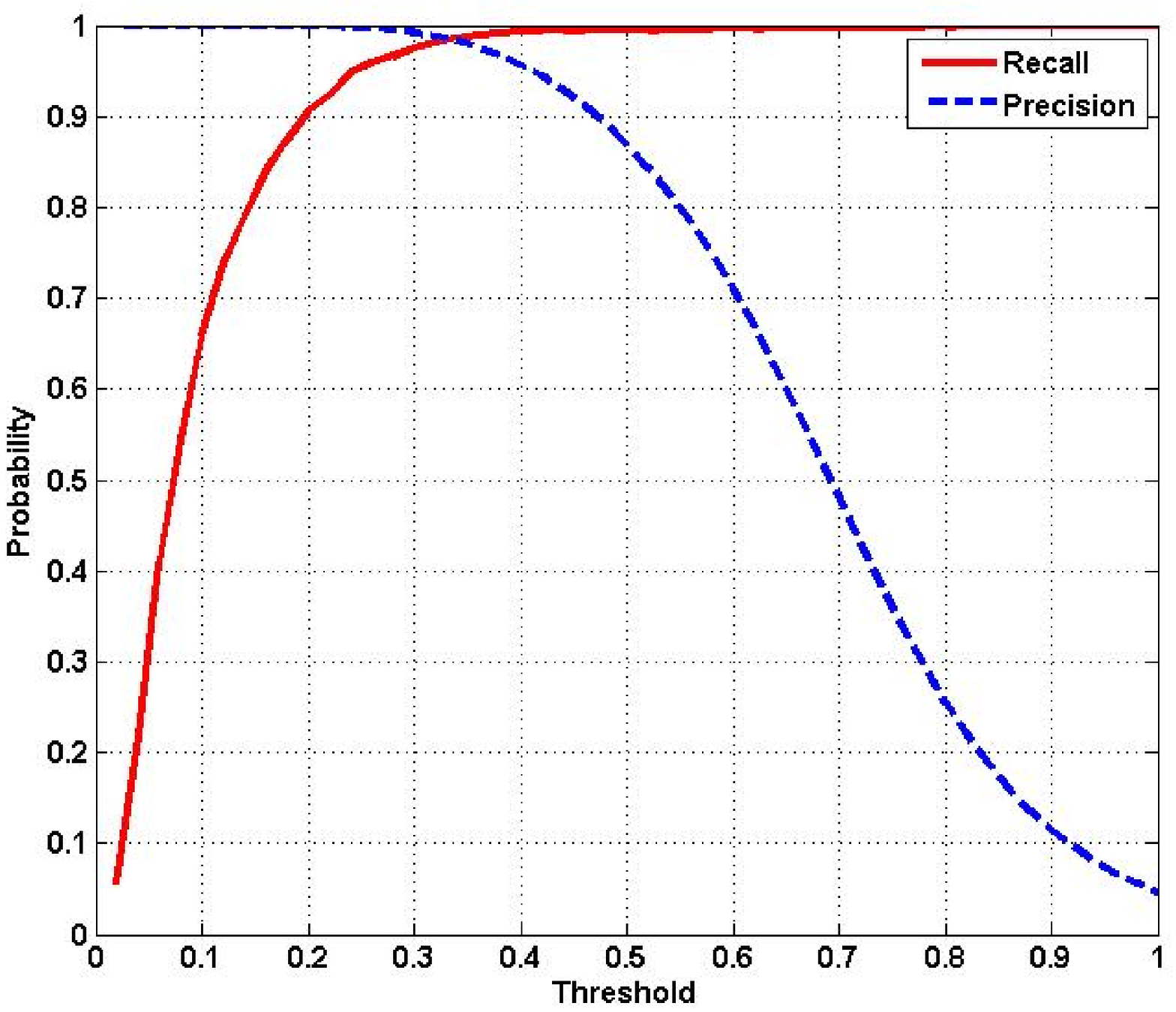}}}
  \caption{ The recall and precision versus thresholds: (a) Affine transformation; (b) Flipping; (c) Letter-box; (d) Logo substitution}\label{inter_compare}
\end{center}
\end{figure*}
\begin{figure*}[htb!]
\begin{center}
  \subfigure[] {\scalebox{0.3}{\includegraphics[width=\textwidth,height=4.2in]{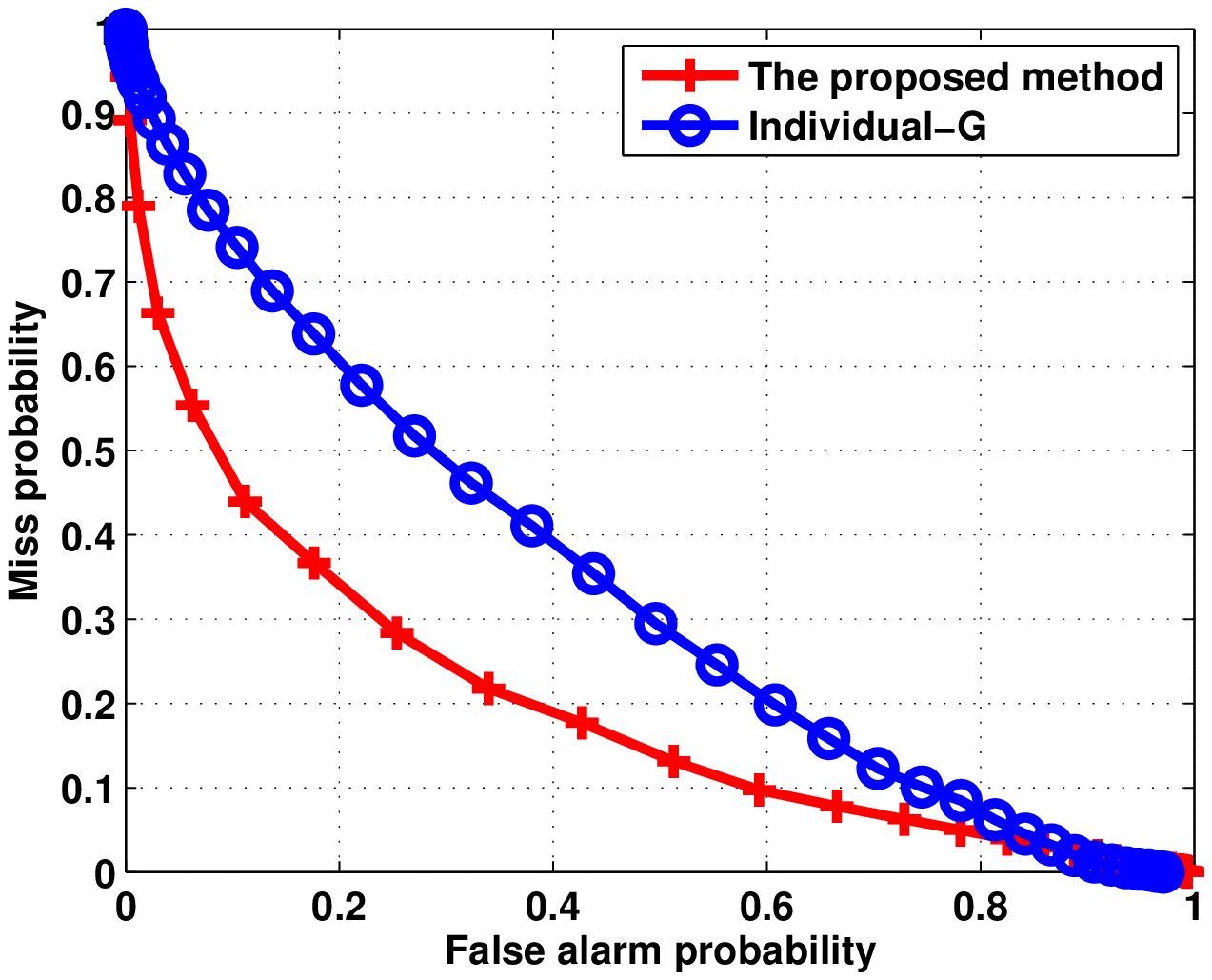}}}
  \subfigure[] {\scalebox{0.3}{\includegraphics[width=\textwidth,height=4.2in]{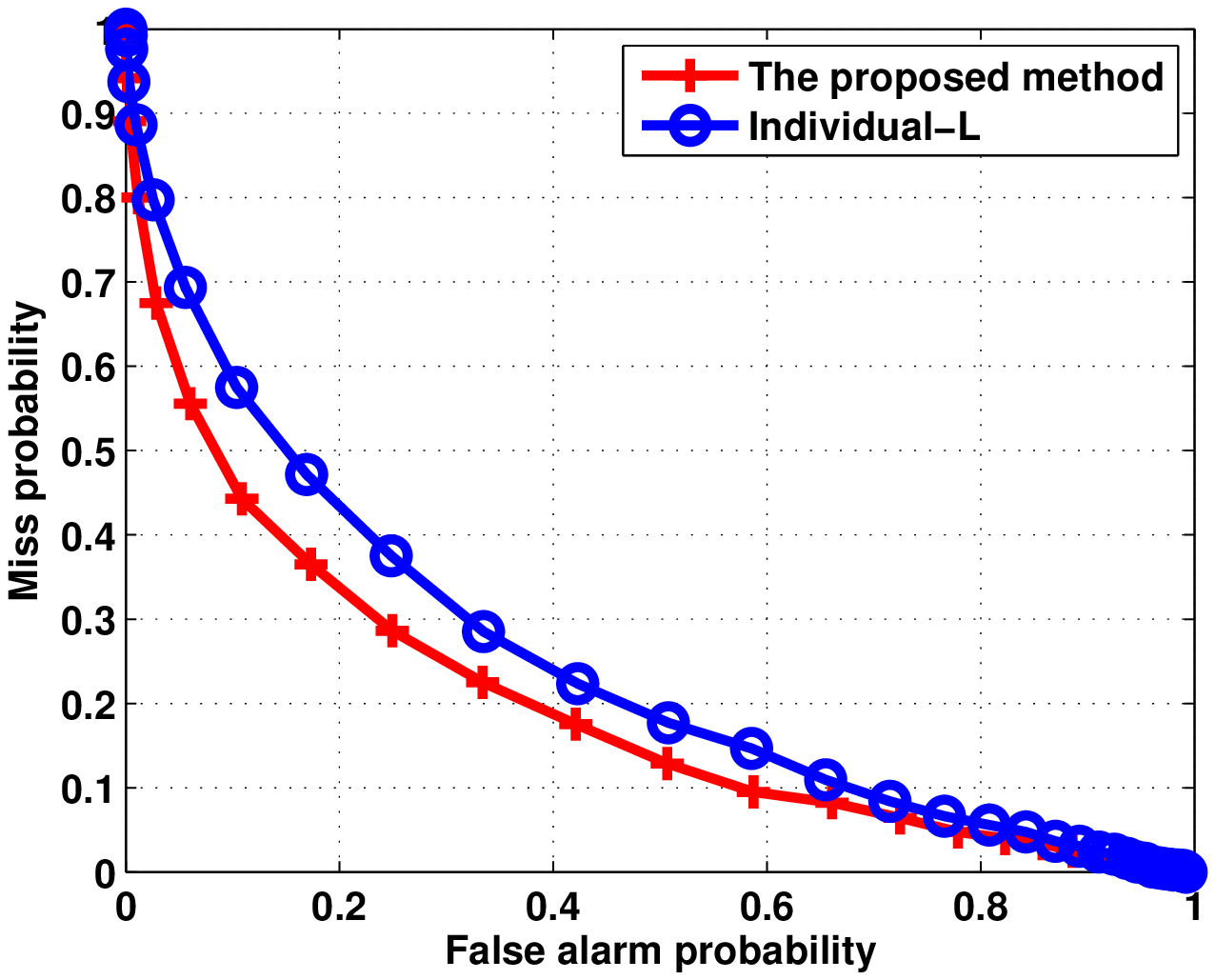}}}
   \subfigure[] {\scalebox{0.3}{\includegraphics[width=\textwidth,height=4.2in]{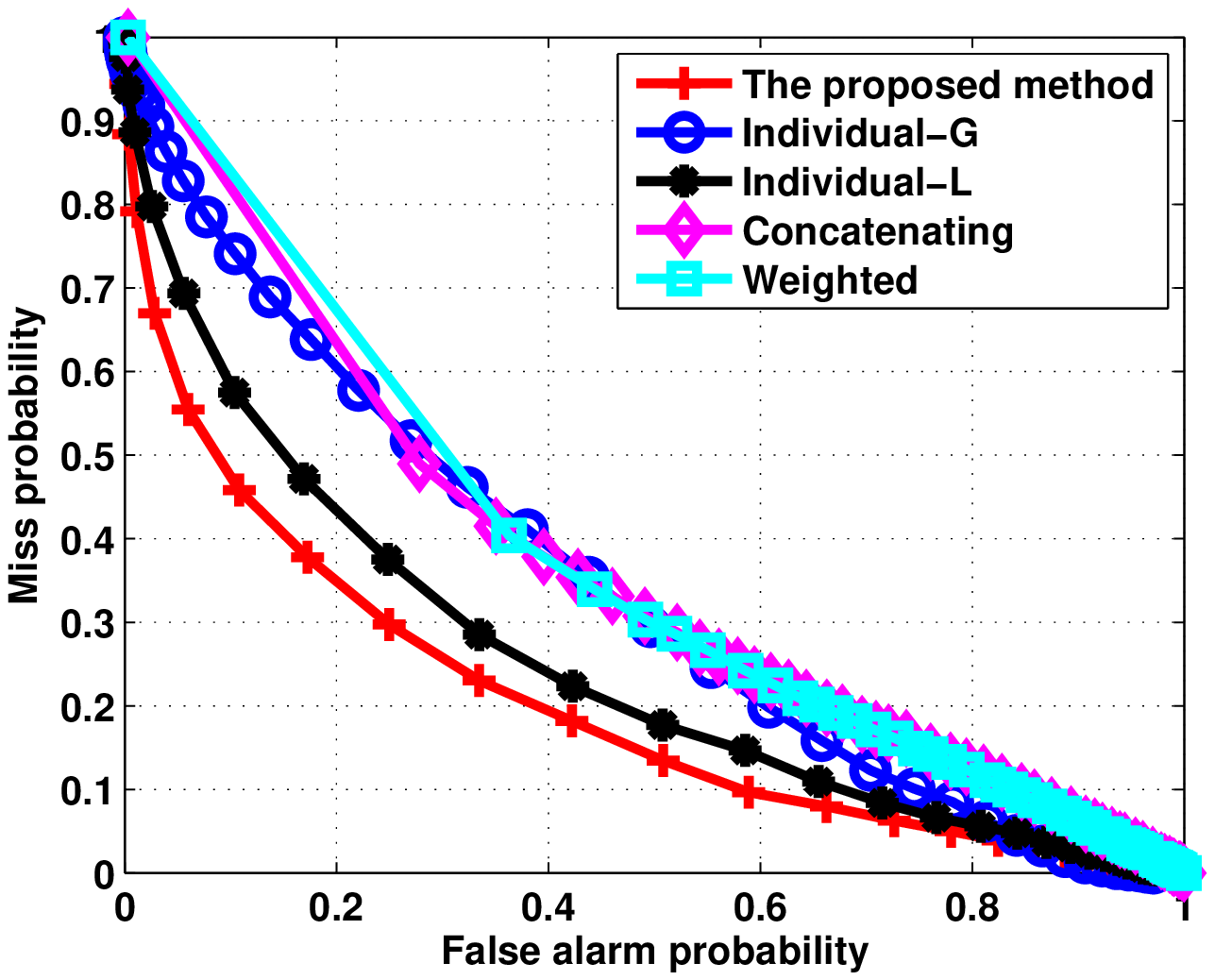}}}
  \caption{ The ROC curve (linear) of comprehensive feature versus the individual feature: (a) The proposed method vs Global feature; (b) The proposed method vs Local feature; (c) The proposed method vs Global/Local/Concatenating feature/Weighted feature}\label{inter_compare}
\end{center}
\end{figure*}

Assuming that $w$ and $v$ yield i.i.d. with Gaussian noise. because the distances between pairs of fingerprint vectors of video contents are independent of each other. Therefore, $w$ and $v$ can be modeled as
\begin{equation}
P(w) \sim N(\mu_w ,\sigma_w ) = \frac{1}{{\sqrt {2\pi } \sigma_w }}{e^{ - \frac{{{{(w - \mu_w )}^2}}}{{2{\sigma_w ^2}}}}}
\end{equation}
\begin{equation}
P(v) \sim N(\mu_v ,\sigma_v ) = \frac{1}{{\sqrt {2\pi } \sigma_v }}{e^{ - \frac{{{{(v - \mu_v )}^2}}}{{2{\sigma_v ^2}}}}}
\end{equation}

To prove the preceding distribution model assumption, we conducted an experiment in the video database under different modifications. Figs. 8 (a) and (b) show the
 values of $v$ and $w$, respectively. The bar of $v$ and $w$ are depicted in Figs. 8 (c) and (d), and the assumption of normal distribution approximatively fits the real data.

According to the model assumptions, we plot the fitting curves in Fig. 9. Obviously, the intersection of these two fitting curves located between 0.3 and 0.4 is a
good choice of the threshold. In these experiments, we set $\tau$=0.32. The plot of recall and precision curves versus thresholds is a way to validate the threshold
 selection discussed. Fig. 10 shows these two curves under certain modifications. The value of the intersection of recall and precision curves,
  which obtained both a high recall and precision probabilities, is approximately the same as the one we obtained using fitting curves.

\subsection{Comprehensive Feature vs Single and Other Fusion Features}
Compared with the single-mode feature, the comprehensive feature is more effective under combined modification in video fingerprinting system. To show the advantages of the
 comprehensive feature, we also generated video fingerprints using two main single-mode features (global and local features), which are normalized 64-bin histograms and
 SURF points, respectively, and then concatenate them to obtain a new feature. We also compared the proposed method with the work in [18], which is a multiple feature fusion-based method.
  They first construct an affine matrix for each feature, and then minimize the distance of hash codes using the affine matrix in each feature. Finally, they sum all
   minimized object functions of different features by weights to optimize the final hashes. We conducted the experiments using its idea without the learning process.
    The comparison of performances between the proposed method and the others under combined modifications (crop+flipping+insertion of patterns) is shown in Fig. 11.
     Figs. 11(a) and (b) show the ROC curves (linear) of the proposed method versus single features, while Fig. 11(c) shows the one between the proposed method and the
      other feature fusion strategies. The performance of the proposed method is better than that of the other methods. Thus, we can conclude that the comprehensive feature has improved the performance well.

\subsection{Matching Performance}
In the proposed scheme, we present a matching strategy based on video core tensor. In this study, we designed an experiment to evaluate the effectiveness of the
proposed strategy. The goal of the pre-matching in the proposed strategy is to narrow the searching range, and the existing matching method can be applied in
 further matching. Therefore, we only take the exhaustive searching for example to show the improvement of the proposed strategy. Eleven different modifications
 were first applied on the video database. Each video in the database was taken as a query. We then tried to find a match in the database with and without the
 proposed strategy in the exhaustive searching. Fig. 12 shows the time used during matching (using a computer with AMD FX-8300 8-core processor, 8 G RAM).
 The adjustment factor and threshold used in the experiment were 0.3 and 0.32, respectively. The matching time is decreased with the proposed
 strategy.
\begin{figure}[htbp] \centering
\includegraphics [width=3in]{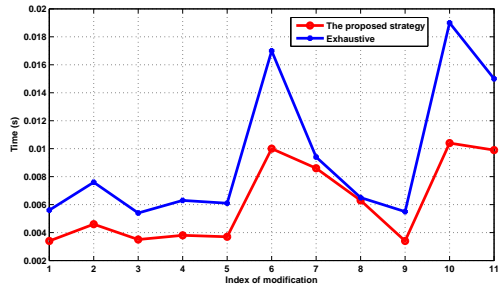}
\caption{The performance of matching}
\label{corticalarchitecturefig}
\end{figure}


\section{Conclusions}
Each feature of a video is not completely independent, and multiple vide features have temporally associated co-occurrence characteristics.
In this study, a novel robust video fingerprinting scheme was proposed to generate the comprehensive feature, which contains the assistance and consensus among different features, to make full use of multiple features.
Compared with the state-of-the-art methods, the proposed method not only fuses multiple features but also captures the intra- and inter-feature
consensus intuitively. The proposed method has good robust and discrimination, especially under the combined modifications.

In this study, we focus on exploiting the assistance and consensus among different features of video fingerprinting system. Which features are selected as the inputs of
the proposed framework is also important in the performance of the entire system because a bad feature or mutually exclusive features degrade the entire performance.
 Thus, the input feature selection is an issue that we will investigate in the future.


\begin{thebibliography}{1}

\bibitem{IEEEhowto:kopka}
H. Shen, X. Zhou, and Z. Huang et al., \emph{UQLIPS: Areal-time near-duplicate video clip detection system}, Proc. of the 33rd int. conf. on Very Large Data Bases, VLDB Endowment, pp. 1374-1377, 2007.

\bibitem{IEEEhowto:kopka}
X. Wu, A. G. Hauptmann, and C. W. Ngo, \emph{Practical elimination of near-duplicates from web video search}, Proc. ACM International Conference on Multimedia, Augsburg, Germany, pp. 218-227, 2007.

\bibitem{IEEEhowto:kopka}
A. Hampapur,  K. H. Hyun, and R. M. Bolle, \emph{Comparison of sequence matching techniques for video copy detection}, Proc. SPIE, Storage and Retrieval for Media Databases, San Jose, CA, USA, pp. 194-201, 2002.

\bibitem{IEEEhowto:kopka}
C. Kim, and B. Vasudev, \emph{Spatiotemporal sequence matching for efficient video copy detection}, IEEE Trans. Circuits Syst. Video Technol., vol. 15, no. 1, pp. 127-132, 2005.

\bibitem{IEEEhowto:kopka}
C. Li, F.W.M. Stentiford, \emph{Video sequence matching based on temporal ordinal measurement}, Pattern Recognition Letters, vol. 41, no. 29, pp. 1824-1831, 2008.

\bibitem{IEEEhowto:kopka}
B. Coskun, B. Sankur, and N. Memon, \emph{Spatio-temporal transformbased video hashing}, IEEE Trans. Multimedia, vol. 8, no. 6, pp. 1190-1208, 2006.

\bibitem{IEEEhowto:kopka}
O. Cirakman, B. Gunsel, and N. Serap et al., \emph{Content-based copy detection by a subspace learning
based video fingerprinting scheme}, Multimedia Tools and Applications, vol. 71, no, 3, pp 1381-1409, 2014

\bibitem{IEEEhowto:kopka}
Z. K. Wei, Y. Zhao, and C. Zhu et al. \emph{Frame fusion for video copy detection}, IEEE Trans. on Circuits and Systems for Video Technology, vol. 21, no. 1, pp. 15-28, 2011.

\bibitem{IEEEhowto:kopka}
A. Joly, O. Buisson, and C. Frelicot, \emph{Content-based copy detection using distortion-based probabilistic similarity search}, IEEE Transactions on Multimedia, vol. 9, no. 2, pp. 293-306, 2007.

\bibitem{IEEEhowto:kopka}
K. Mikolajczyk and C. Schmid, \emph{A performance evaluation of local descriptors}, IEEE Trans. Pattern Anal. Mach. Intell., vol. 27, no. 10, pp. 1615-1630, 2005.

\bibitem{IEEEhowto:kopka}
S. Lee and C. D. Yoo,\emph{ Robust video fingerprinting for content-basedvideo identification}, IEEE Trans. Circuits Syst. Video Technol., vol. 18, no. 7, pp. 983-988, 2008.

\bibitem{IEEEhowto:kopka}
G. Yang, N. Chen and Q. Jiang, \emph{A robust hashing algorithm based on SURF for video copy detection}, Computers \& Security, vol. 31, no. 1, pp. 33-39, 2012.

\bibitem{IEEEhowto:kopka}
M. Li and V. Monga, \emph{Compact video fingerprinting via structural graphical models}, IEEE Transactions on Information Forensics and Security, vol.  8, no. 11, pp. 1709-1721, 2013.

\bibitem{IEEEhowto:kopka}
A. S. Divya, D. G. Wiselin, \emph{ Video copy detection using spatio-temporal and texture features}, International Journal of Emerging Technology and AdvancedEngineering, vol. 2, pp. 293-296, 2012.

\bibitem{IEEEhowto:kopka}
X. C. Liu, J. D. Sun, and J. Liu, \emph{Visual attention based temporally weighting method for video hashing}, IEEE Signal Processing Letters, vol. 20, no. 12, pp. 1253-1256, 2013.

\bibitem{IEEEhowto:kopka}
X. S. Nie, J. D. Sun, and Z. H. Xing et al., \emph{Video fingerprinting based on graph model}, Multimedia Tools and Applications, vol. 69, no. 2, pp. 429-442, 2014.

\bibitem{IEEEhowto:kopka}
X. S. Nie, J. Liu, and J. D. Sun et. al., \emph{Robust video hashing based on representative-dispersive frames}, Science China: Information Sciences, vol. 56, no. 6,  pp. 1-11, 2013.

\bibitem{IEEEhowto:kopka}
J. K. Song, Y. Yang, Z. Huang, and H. T. Shen et al.,
\emph{Effective Multiple Feature Hashing for Large-Scale
Near-Duplicate Video Retrieval}, IEEE Transactions on Multimedia, vol. 15, no. 8, pp. 1997-2008, 2013.

\bibitem{IEEEhowto:kopka}
M. L. Jiang, Y. H. Tian, and T. J. Huang, \emph{Video copy detection using a soft cascade of
multimodal features}, In Proceedings of the International Conference on Multimedia and Expo. pp. 374¨C379, 2012

\bibitem{IEEEhowto:kopka}
Y. N. Li, L. T. Mou, and M. L. Jiang et al., \emph{Copy detection with
visual-audio feature fusion and sequential pyramidmatching}, PKU-INM@ TRECVid 2010

\bibitem{IEEEhowto:kopka}
Y. H. Tian, M. L. Jiang, and L. T. Mou et al., \emph{A multimodal video
copy detection approach with sequential pyramid matching}, In Proceedings of the International Conference
on Image Processing, pp. 3629¨C3632, 2011

\bibitem{IEEEhowto:kopka}
F. Wu, Y. N. Liu, and Y. T. Zhuang, \emph{Tensor-Based Transductive Learning for Multimodality Video Semantic Concept Detection}, IEEE Transactions on Multimedia, vol. 11, no. 5, pp. 868-878, 2009.

\bibitem{IEEEhowto:kopka}
M. Li and V. Monga, \emph{Desynchronization resilient video fingerprinting via randomized low-rank tensor approximations}, Proc. IEEE Int. Workshop Multimedia Signal Process., Hangzhou, China, pp.1-6, 2011.

\bibitem{IEEEhowto:kopka}
M. Li and V. Monga, \emph{Robust Video Hashing via Multilinear Subspace Projections}, IEEE Transactions on Image Processing, vol. 21, no. 10, pp. 4397-4409, 2012.


\bibitem{IEEEhowto:kopka}
T. G. Kolda, B. W. Bader, \emph{Tensor Decompositions and Applications}, SIAM Review, vol. 51, no. 3 pp. 455-500, 2009.



\bibitem{IEEEhowto:kopka}
B. H. Tuytelaars, T. G. LV, \emph{SURF: speeded up robust features}, Computer Vision and Image Understanding, vol. 110, no. 3, pp.346-359, 2008.

\bibitem{IEEEhowto:kopka}
E. Ceulemans, H. A. L. Kiers, \emph{Selecting among three-mode principal component models of different types and
complexities: A numerical convex hull based method}, British Journal of Mathematical and Statistical Psychology, vol. 59, no. 5, pp.113-150, 2006.

\bibitem{IEEEhowto:kopka}
M. M{\o}rup, L. K. Hansen, \emph{Automatic relevance determination for
multi-way models}, Journal of Chemometrics, Special Issue: In Honor of Professor Richard A. Harshman, vol. 23, no. 7-8, pp. 352-363, 2009.

\bibitem{IEEEhowto:kopka}
M. M{\o}rup, \emph{Decomposition methods for unsupervised learning}. PhD
Thesis, Technical University of Denmark, 2008.

\bibitem{IEEEhowto:kopka}
M. Morten, \emph{Applications of tensor (multiway array) factorizations and decompositions in data mining}, Wiley Interdisciplinary Reviews: Data Mining and Knowledge Discovery, vol. 1, no. 1, pp. 24-40, 2011.

\bibitem{IEEEhowto:kopka}
G. Awad,P. Over, and W. Kraaij, \emph{Content-based video copy detection benchmarking at TRECVID}, ACM Transactions on Information Systems, vol. 32, no. 3, Article 14: 1-40, 2014.

\bibitem{IEEEhowto:kopka}
M. M. Esmaeili, M. Fatourechi, and R. K. Ward, \emph{A Robust and Fast Video Copy Detection System Using Content-Based Fingerprinting}, IEEE Transactions on Information Forensics and Security, vol. 6, no. 1, pp. 213-226, 2011.

\bibitem{IEEEhowto:kopka}
F. Comon, X. Luciani, and A. L. F. D. Almeida, \emph{Tensor decompositions alternating least square and other tales}, J. Chemometrics, vol. 23, no. 7, pp. 393-405, 2009.

\bibitem{IEEEhowto:kopka}
B. De Lathauwer, D. Moor, and J. Vandewalle, \emph{On the best rank-1 and rank-(R1,R2, . . . ,RN) approximation of higher-order tensors}, SIAM J. Matrix Anal. Appl., vol 21, pp. 1324-1342, 2000.

\bibitem{IEEEhowto:kopka}
P. M. Kroonenberg, J. De Leeuw, \emph{Principal component analysis of three-mode data by
means of alternating least squares algorithms}, Psychometrika, vol 45, pp. 69-97, 1980.

\bibitem{IEEEhowto:kopka}
C. Xu, D. Tao, C. Xu, \emph{A survey on multi-view learning}, arXiv:1304.5634, 2013.

\bibitem{IEEEhowto:kopka}
S. Dasgupta, M.L. Littman, and D. McAllester. \emph{Pac generalization bounds for co-training},
Advances in neural information processing systems, 1, pp. 375-382, 2002.

\bibitem{IEEEhowto:kopka}
Hansen L. K., Madsen K. H., Lehn S. T., \emph{Adaptive regularization of
noisy linear inverse problems}, In Proceedings of Eusipco 2006, 2006

\end{thebibliography}
\end{document}